\documentclass{technojournal}

\received{April 12th, 2019}
\revised{April XX, 2019}
\accepted{April XX, 2019}

\title[MIR-Vehicle: Cost-Effective Research Platform for Autonomous Vehicle Applications]{MIR-Vehicle: Cost-Effective Research Platform for Autonomous Vehicle Applications}

\correspinfo[Professor Jaerock Kwon]{jrkwon@umich.edu}

\author[a]{Ahmed Abdelhamed}{\thanks{E-mail: \email{abde5186@kettering.edu}}}
\author[a]{Balakrishna Yadav Peddagolla}{\thanks{E-mail: \email{pedd4463@kettering.edu}}}
\author[a]{Girma Tewolde}{\thanks{E-mail: \email{gtewolde@kettering.edu}}}
\author[b]{Jaerock Kwon \corresp}{}
\shortauthor{A. Abdelhamed, B. Peddagolla, G. Tewolde and J. Kwon} 

\affiliation[a]{The Department of Electrical and Computer Engineering, Kettering University, 1700 University Avenue, Flint, MI 48504-6214, USA}
\affiliation[b]{The Department of Electrical and Computer Engineering, the University of Michigan-Dearborn, 4901 Evergreen Road, Dearborn, MI 48128, USA}

\usepackage{graphicx}
\usepackage{tabu,hhline,multirow}

\keywords{Intelligent Robots; Mobile Robots; Robot Design; Robotics in Intelligent Vehicle and Highway Systems; Mechatronic Systems;}

\newcommand{\ba}{\begin{align}}
\newcommand{\ea}{\end{align}}

\begin{document}

\maketitle

\begin{abstract}
This paper illustrates the MIR (Mobile Intelligent Robotics) Vehicle: a feasible option of transforming an electric ride-on-car into a modular Graphics Processing Unit (GPU) powered autonomous platform equipped with the capability that supports test and deployment of various intelligent autonomous vehicles algorithms. To use a platform for research, two components must be provided: perception and control. The sensors such as incremental encoders, an Inertial Measurement Unit (IMU), a camera, and a LIght Detection And Ranging (LIDAR) must be able to be installed on the platform to add the capability of environmental perception. A microcontroller-powered control box is designed to properly respond to the environmental changes by regulating drive and steering motors. This drive-by-wire capability is controlled by a GPU powered laptop computer where high-level perception algorithms are processed and complex actions are generated by various methods including behavior cloning using deep neural networks. The main goal of this paper is to provide an adequate and comprehensive approach for fabricating a cost-effective platform that would contribute to the research quality from the wider community. The proposed platform is to use a modular and hierarchical software architecture where the lower and simpler motor controls are taken care of by microcontroller programs, and the higher and complex algorithms are processed by a GPU powered laptop computer. The platform uses the Robot Operating System (ROS) as middleware to maintain the modularity of the perceptions and decision-making modules. It is expected that the level three and above autonomous vehicle systems and Advanced Driver Assistance Systems (ADAS) can be tested on and deployed to the platform with a decent real-time system behavior due to the capabilities and affordability of the proposed platform.
\end{abstract}

\section{Introduction}
Autonomous Vehicles (AV) have been realized by rapid advances in technology including sensors and computing platforms with artificial intelligence, and more attention has been drawn in research communities to developing the systematic testing and evaluating of complex perception and control algorithms. A full-scale drive-by-wire vehicle, however, is often cost prohibitive for research labs, and may not even be necessary. To conduct research on perception and control for autonomous vehicles, all that is required is a platform controlled by electronic systems equipped with sensors. We believe that the size of the platform does not matter in conducting such research in many cases. Thus, we propose the MIR (Mobile Intelligent Robotics) Vehicle that is a small but scalable research platform for autonomous vehicle applications. This is a feasible option since the proposed platform is an electric ride-on-car equipped with various sensors that are connected to a Graphics Processing Unit (GPU) powered computer. There even exist advantages of using a small-scale vehicle, such as significantly lower safety risk in testing and analyzing new algorithms compared with doing so on a full-size vehicle. To be a viable platform for research, two components must be provided: perception and control. The sensors such as incremental encoders, an Inertial Measurement Unit (IMU), a camera, and a LIght Detection And Ranging (LIDAR) must be able to be installed on the platform to add the capability of environmental perception. Most of the vehicles do not provide accessible electronic controls over the basic vehicle functions as they are mainly designed for human-machine interface rather than a machine-machine interface. Therefore, some work is needed to be done to make a remote control ride-on-car into a drive-by-wire one. First, a motor encoder is attached to each of the steering and driving gearboxes. Then, a microcontroller-powered control box is prepared to properly regulate driving and steering motors. This drive-by-wire capability is controlled by a GPU powered laptop computer where high-level perception algorithms are processed and complex actions are generated by various methods including behavior cloning using deep neural networks. 
The main goal of this paper is to provide an adequate and comprehensive approach for building a cost-effective platform that would contribute to the research quality for the wider community. The proposed platform is to use a modular and hierarchical software architecture where the lower and simpler motor controls are taken care of by microcontroller programs, and the higher and complex algorithms will be processed by a GPU powered laptop computer. The platform uses the Robot Operating System (ROS) \citep{quigley_ros:_2009} as middleware to maintain the modularity and distributed computing of the perceptions and decision-making modules. We believe that the level three and above autonomous driving (AD) systems and Advanced Driver Assistance Systems (ADAS) can be tested on and deployed to the platform with a decent real-time system behavior due to the capabilities and affordability of the proposed platform. 

\begin{figure}
\centering
\includegraphics[width=0.5\textwidth]{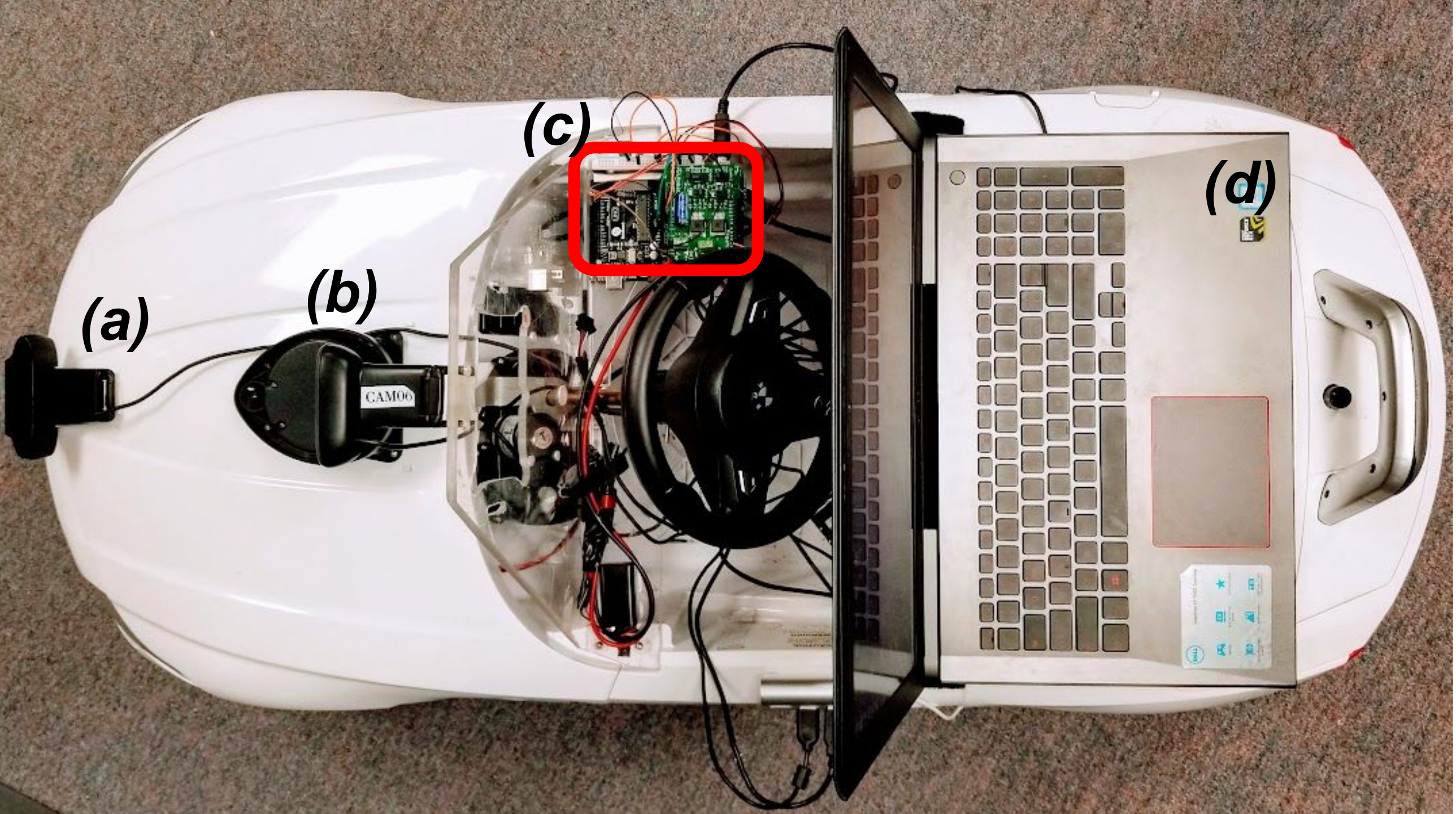}
\includegraphics[width=0.42\textwidth]{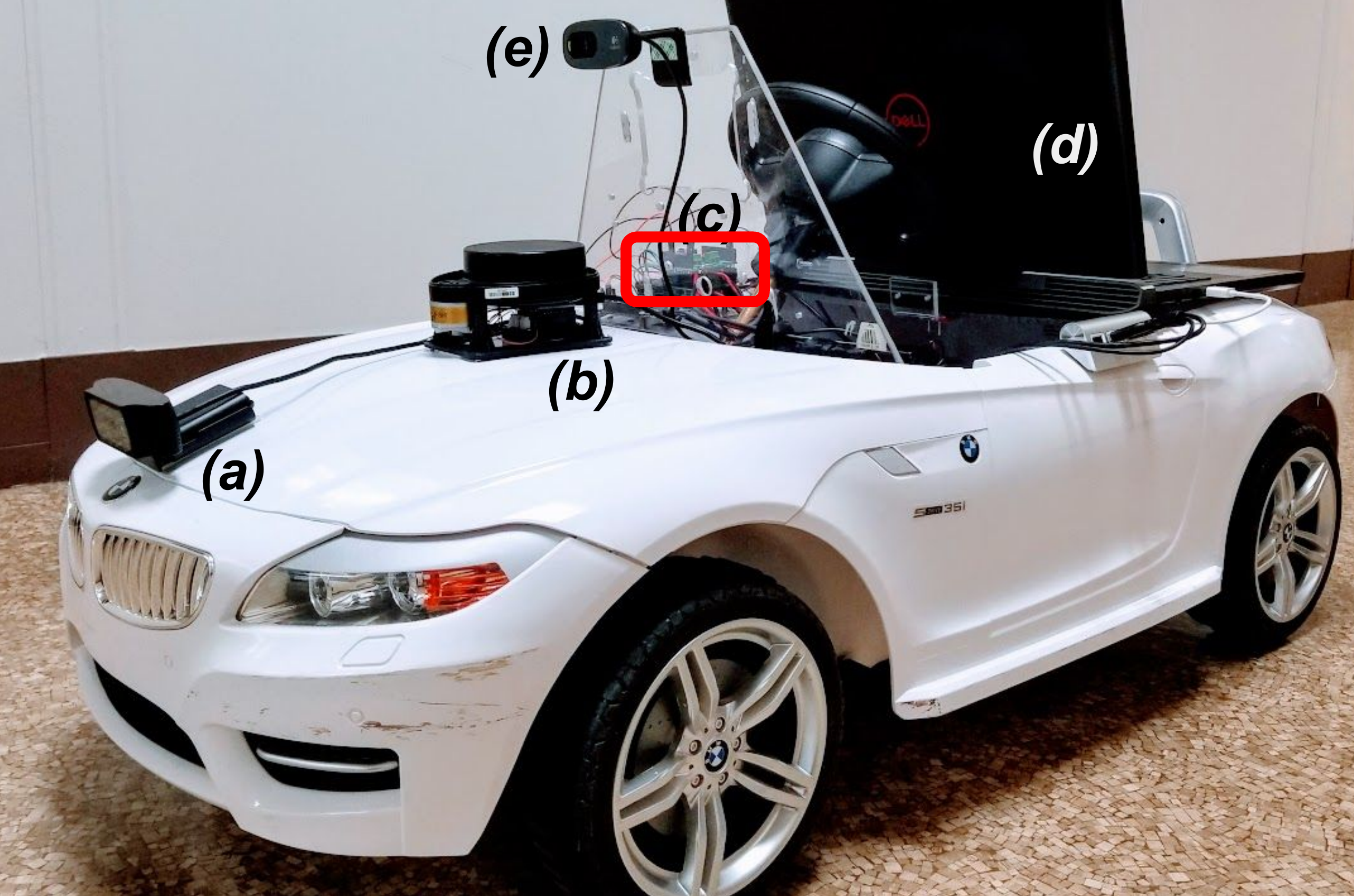}
\caption{MIR-Vehicle. Top: The top view of the MIR Vehicle. Bottom: The profile view. (a) is a camera and (b) is a LIDAR for data acquisition. (c) is the motor control module. (d) is a laptop computer. (e) is a camera to take first-person-view videos for the platform.
}
\label{fig-mir-vehicle-view}
\end{figure}

This paper describes the design processes of transforming a ride-on-car into an autonomous testing platform. Fig.~\ref{fig-mir-vehicle-view} shows the overview of the MIR Vehicle. The processes include adding various sensors like cameras, LIDAR’s, 9 Degree of Freedom (DOF)  IMU’s, and 3D depth cameras and carrying out the modifications of the gearboxes of the vehicle to add incremental encoders. These sensors contribute to applying various autonomous intelligent algorithms such as obstacles detection and classification, lane keeping assist, mapping a vehicle's environment with the capability of applying the Simultaneous Localization and Mapping (SLAM). This paper, also, comprehensively discusses the configuration and setup of each of the above-mentioned sensors along with the ROS-based software architecture.

\section{Related Works}
Most of the major established automobile companies, as well as the relatively recent technology players such as Tesla and Waymo, are working on the development of autonomous vehicles in some ways. According to a report from \citet{cb-insigths-research_46_2018}, Aptiv conducted more than 5,000 self-driving taxi rides in Las Vegas in 2018; Autoliv aimed to commercialize driver assistance technology by 2019; Ford acquired Argo and plans to release autonomous vehicles by 2021; GM Cruise Automation has been working on commercializing ride-sharing service through autonomous Chevy Bolts. As autonomous vehicle applications make the transition from research laboratories to the real world, many vehicle platforms have been proposed and tested. 

Much effort has been made in the following three different approaches. The first one is to modify a real vehicle. A research team at Carnegie Mellon University proposed a research platform to meet the requirements of general autonomous driving \citep{wei_towards_2013}. This minimally modified platform in its appearance improved social acceptance while demonstrating the viability of the product. Another design of an autonomous car was introduced from \citet{jo_development_2014, jo_development_2015}. They developed an autonomous car with distributed system architecture to efficiently process complex algorithms and deal with the heterogeneity of the components. These designs are, however, for a particular vehicle and not easy to be replicated due to in-house components and no availability of the control software. Therefore, the design and development could hardly be used by other research groups not to mention the prohibitive cost of building the system. The second approach is to use a simulation environment. In the real world, it costs a lot money and it is a time-consuming process to develop and test algorithms for autonomous vehicles. AirSim is a visual and physical simulation for autonomous vehicles \citep{shah_airsim:_2018}. This simulator was built on the Unreal Engine for realistic simulation. CARLA is an urban driving simulator to support development and validation of an autonomous driving system \citep{dosovitskiy_carla:_2017}. These attempts to use simulators are useful but Hardware-in-the-Loop (HIL) is still necessary to test certain perception and control algorithms. In the third approach, there have been efforts to develop cost-effective autonomous vehicle research platforms. We believe that this approach is more viable and worth for most research groups since these designs are replicable and affordable. The following are recent research activities that employ this approach.

A ROS-based 1/10th scale Remote Control (RC) car, Cherry Autonomous Race Car, was built using Traxxas and NVIDIA’s Jetson TX1 with a cost of \$1,800 \citep{tobias_daniel_2017}. The goal of this project was to convert an RC car into an autonomous vehicle by implementing Convolution Neural Networks (CNN) on NVIDIA’s Jetson TX1. This approach prohibits us from using a more powerful computing platform and adding more sensors since the RC car platform is too small to hold a bigger computer and additional sensors. In addition, Jetson TX1 was developed to give a GPU powered machine learning libraries for a small embedded system. There is no reason to limit us to use an embedded platform if we have another viable choice. 

A go-kart sized low-cost autonomous vehicle was created by \citet{ollukaren_low-cost_2014}. In this design, the main computer was a Raspberry Pi board that is a small single board computer. As proof of concept, they programmed the computer to detect and follow a sheet of red paper. It is not a reasonable choice of using the limited performance Raspberry Pi single-board computer with a go-kart sized platform. 

A ROS-based home-made mobile robotic platform was proposed by \citet{gomez_clara_ros-based_2015}. This work was focused on developing a two-wheel differential driving robotic platform. The platform was equipped with a low-cost microcontroller to process the odometry information, and the Radxa Rock Pro \citep{radxa-rock-pro_radxa_2014} was used to process a 3D depth camera and perform a SLAM algorithm. Due to the limited size of the platform, a small single-board computer without GPU powered was used, and this limits the system from using the state-of-the-art machine learning libraries.
An Autonomous Vehicle Research Platform (AVRP) was developed and offered researchers an autonomous ground vehicle testing platform \citep{walling_design_2017}. The scope of the work was to develop the operational specifications that can operate at level four autonomy. The design specs, however, for power and communication buses were not discussed in details.

A low-cost ROS-based platform, RoboMuse 4.0, was introduced by \citet{shukla_development_2017}. The aim of the project was to create an educational platform to be used by researchers. The platform was equipped with on-wheel incremental encoders and a 3D depth camera. It used a microcontroller to interface with a GPU powered laptop. SLAM and object recognition were implemented using the platform. This paper reported their implementation but no details of their design were provided.
The Autonomous Vehicle Control System Kit (AVCS Kit) was proposed in \citet{dang_designing_2017}. They described a control system that was able to convert a ride-on-car into an autonomous vehicle. A lane detection algorithm was implemented to show the feasibility of the proposed system. A fisheye lens camera was used to address the narrow view-angle problem to see side lanes. The cost of the AVCS kit was around \$360 excluding a laptop computer and a SICK LIDAR. In-house control software was used and not publicly available. Thus, we believe that there may be a scalability issue on the design.

An automotive platform for ADAS was developed using an electric go-kart in \citep{alzubi_cost_2018}. The ADAS algorithms were developed within the ROS. To show their system’s feasibility, Lane Keeping Assist (LKA) and Automatic Emergency Braking (AEB) algorithms were also presented. The budget of the proposed system was much higher than the previously discussed platforms.

In our work, we concentrated not only on creating a cost-effective platform for testing autonomous vehicle algorithms, but we also focused on how to utilize the ROS as the middleware for the system to improve the scalability of the system.

\section{Methods}
After analyzing other approaches to build a cost-effective vehicle platform, our design for the cost-effective research platform for autonomous vehicle applications is as follows. (1) The computing platform must not be a single-board computer but a GPU powered laptop that is capable of running state-of-the-art machine learning libraries. (2) The size of the car is big enough to hold various sensor packages and a GPU powered laptop, and small enough to easily carry or to pose any safety risks when using the platform. (3) All software modules, except the low-level device drivers for motors, will be written as ROS nodes, that are processing units performing computation. The hardware design of the proposed platform was mainly inspired by the work of \citet{dang_designing_2017}, but we designed the software package of the platform to be hierarchical and modular with scalability powered by the ROS.

\subsection{The MIR-Vehicle}
\begin{figure}
\centering
\includegraphics[width=0.6\textwidth]{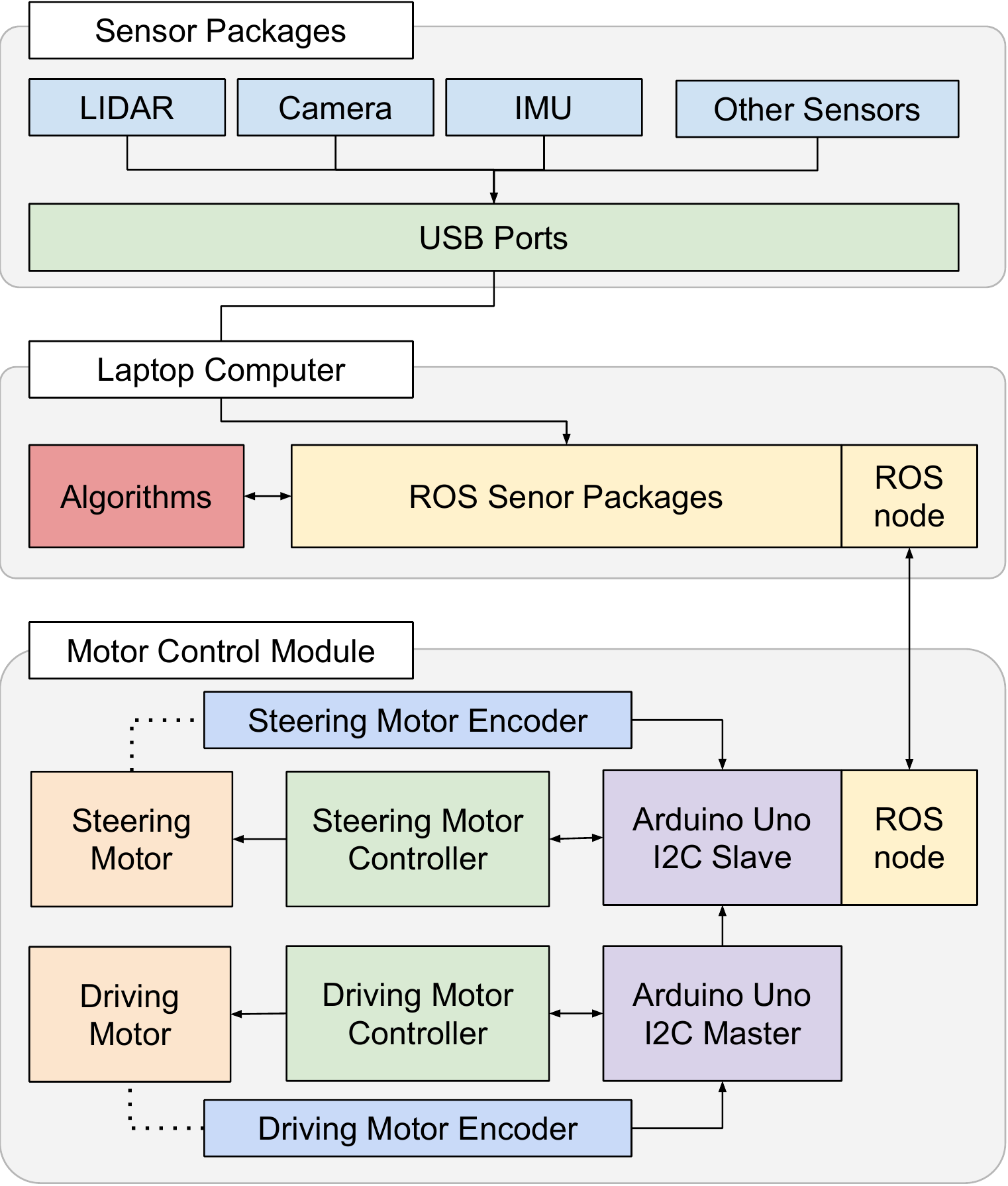}
\caption{The block diagram of the MIR-Vehicle platform. The laptop computer hosts device drivers of the sensor packages and runs a ROS node that communicates with the microcontrollers through the ROS serial communication. The motor control module\textsc{\char13}s only concern is the low-level motor control and it is independent of the main unit where higher level perception and control algorithms are executed. The components in one module are easily replaceable without affecting other modules.}
\label{fig-mir-vehicle-architecture}
\end{figure}

The motor control module takes care of the low-level motor control tasks such as counting a motor encoder’s signal pulses and sending motor driving Pulse Width Modulation (PWM) signals. The laptop does all the high-level computational service, reads the data from sensor packages using ROS nodes, and runs all algorithms created for testing on the platform. A schematic diagram of the architecture of the MIR Vehicle platform is shown in Fig.~\ref{fig-mir-vehicle-architecture}. This research platform supports hierarchical, modular, and scalable design. Each component is discussed in the following sections in more details. The platform is based on ROS compatible packages using Python. The ROS provides a distributed computing environment by which all communication between components in the system is seamlessly possible. The MIR Vehicle platform offers capabilities to develop perception and control algorithms and deploy them to a vehicle model. Sensor packages are connected via a USB port to a laptop computer in which perception and control algorithms execute. The laptop computer hosts the ROS core that is a meta package to aggregate the packages required to use publish and subscribe services. A ROS node runs in the motor control module to receive control messages from the main computer and send the information of encoders to the ROS nodes in the laptop computer. More details will be given in the Software Design Overview section. The motor control unit takes care of motor driving and feedback of the motors using the encoders. Two motor controllers are used to handle two DC motors. Also, two microcontrollers are deployed to the unit to count motor revolutions. One of them is used to execute a ROS node that sends encoder values to and receives commands from the laptop computer.

\subsection{Car Platform}
Several considerations were made in the car platform selection. First, the ride-on-car must be large enough to hold a GPU enabled laptop computer with sensors and small enough to be easily carried. Second, steering and driving must be able to be controlled by PWM signals. Third, there must be enough space where encoders can be installed to measure both the angles of the steering wheel and the speed of the driving motor. Our choice of car is the electric ride-on-car from \citet{bmw-z4-roadster_bmw_2009}. The product dimension is 1.11 x 0.57 x 0.47 meters. The weight is 14.15 kg. The wheel’s diameter is 0.2 meters. The maximum speed is 1.12 m/s. The maximum load is 30 kg. The driving motor is DC 6 V, 25 Watt. The original battery is 6 V and 7 Ah. We replaced the battery with two separate batteries: a 9 V, 2000 mAh battery for the driving motor and a 9 V 600 mAh battery for the steering wheel motor to separate power source for each motor. Using two separate batteries are beneficial since they are lighter than the original battery and easier to recharge.

\subsection{Mechatronics Design}
The transformation process of the ride-on-car into an autonomous ROS based platform requires both mechanical and electrical adjustments. 

\subsubsection{Mechanical Design}
\begin{figure}
\centering
\includegraphics[width=0.48\textwidth]{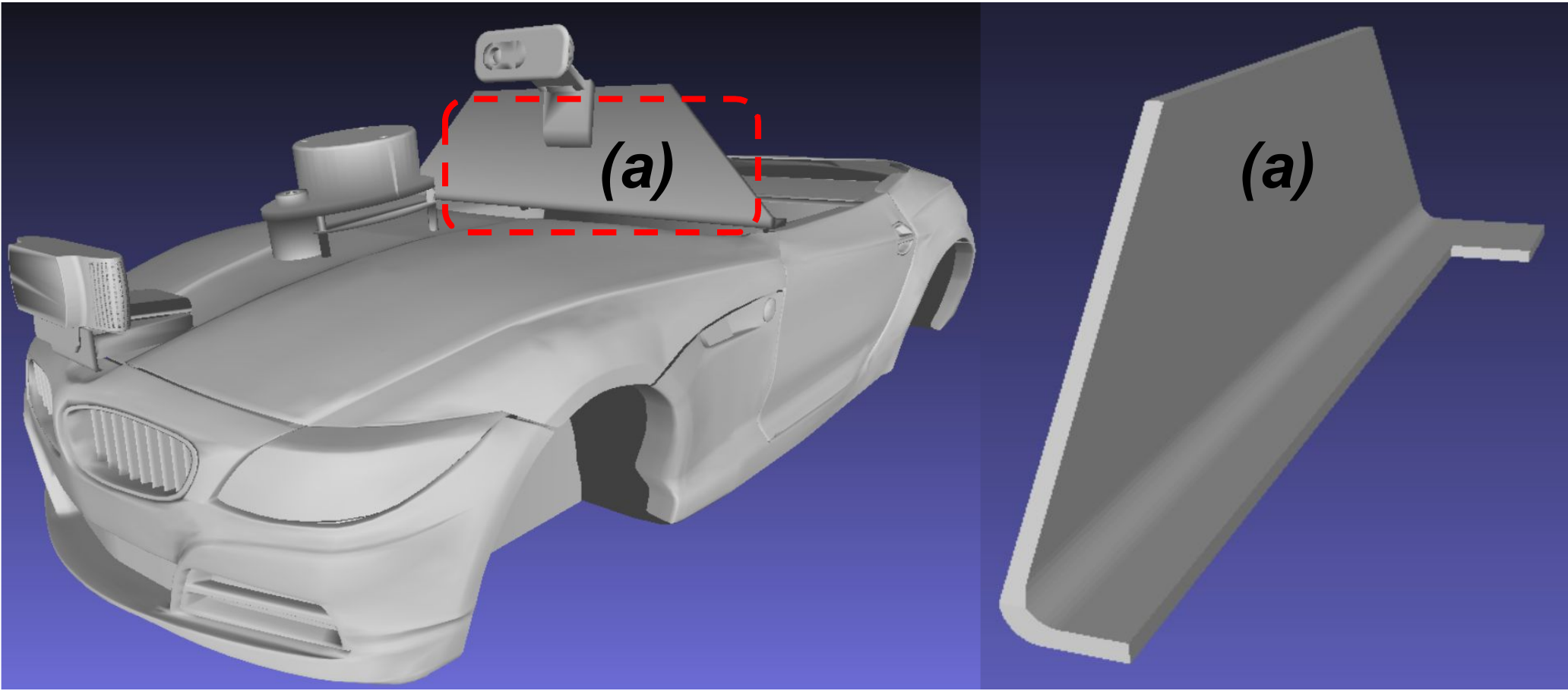}
\includegraphics[width=0.48\textwidth]{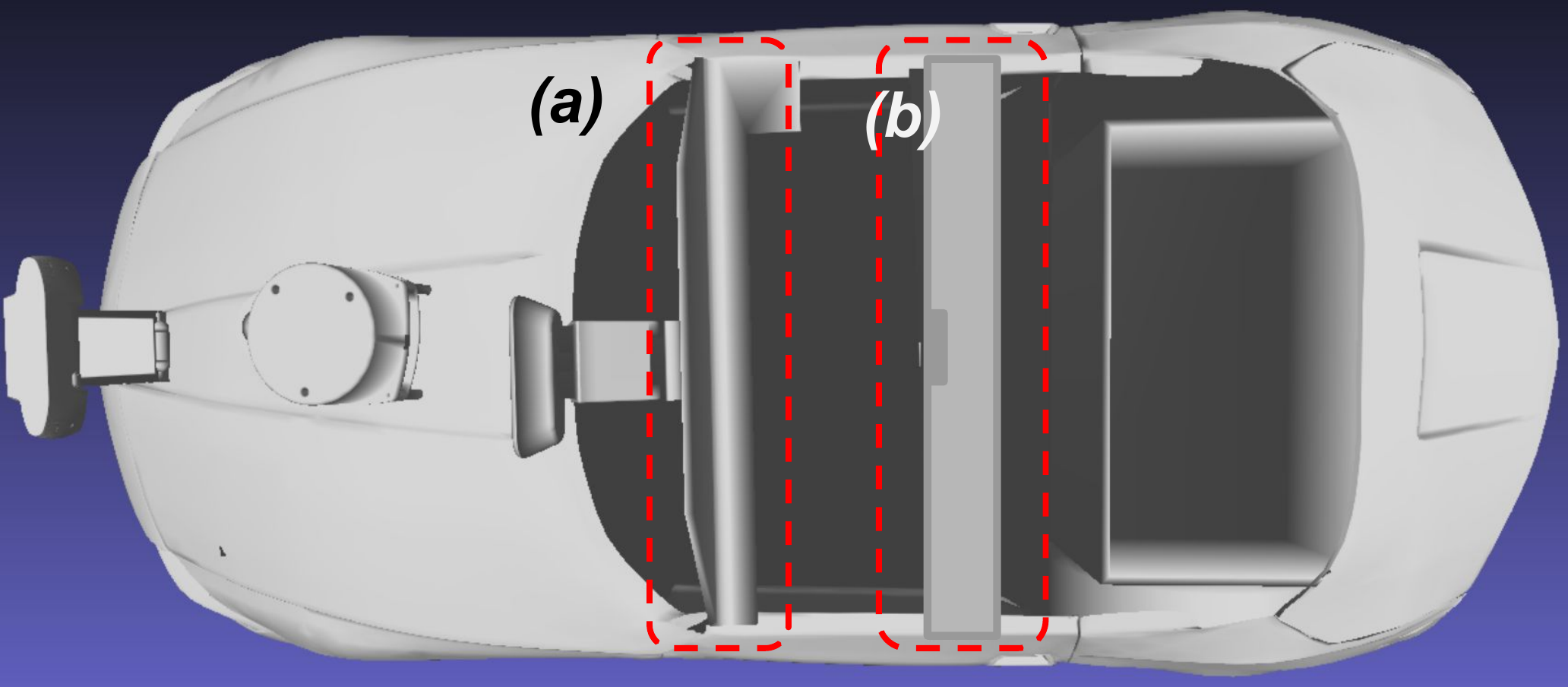}
\caption{The rigid acrylic fixtures. (a) The replacement of the original windshield of the car. It holds sensors as well as the motor control box. (b) The laptop computer holder. The original seat was removed, and a rigid acrylic board was affixed to have a stable position for the laptop computer.}
\label{fig-rigid-acrylic-fixtures}
\end{figure}

We designed a windshield replacement that is used to hold sensors and wiring harness. Additional electric wires from motors to the two motor drivers must be connected. The rigid acrylic fixtures are used to hold sensors as well as the motor control box. The laptop computer holder is also necessary to have a stable position for the laptop computer. See Fig.~\ref{fig-rigid-acrylic-fixtures} for more details. 

\begin{figure}
\centering
\includegraphics[width=0.3\textwidth]{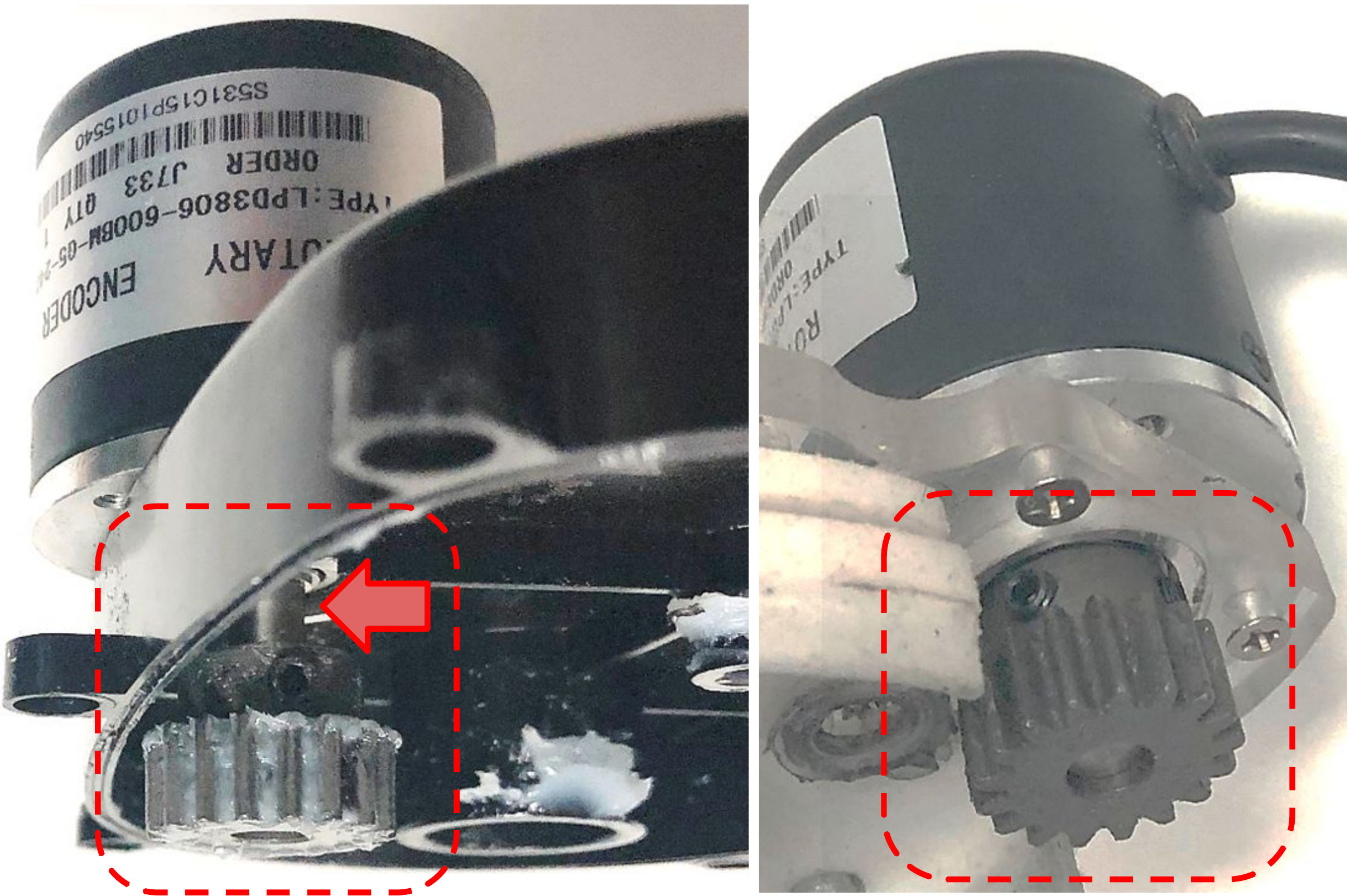}
\includegraphics[width=0.2\textwidth]{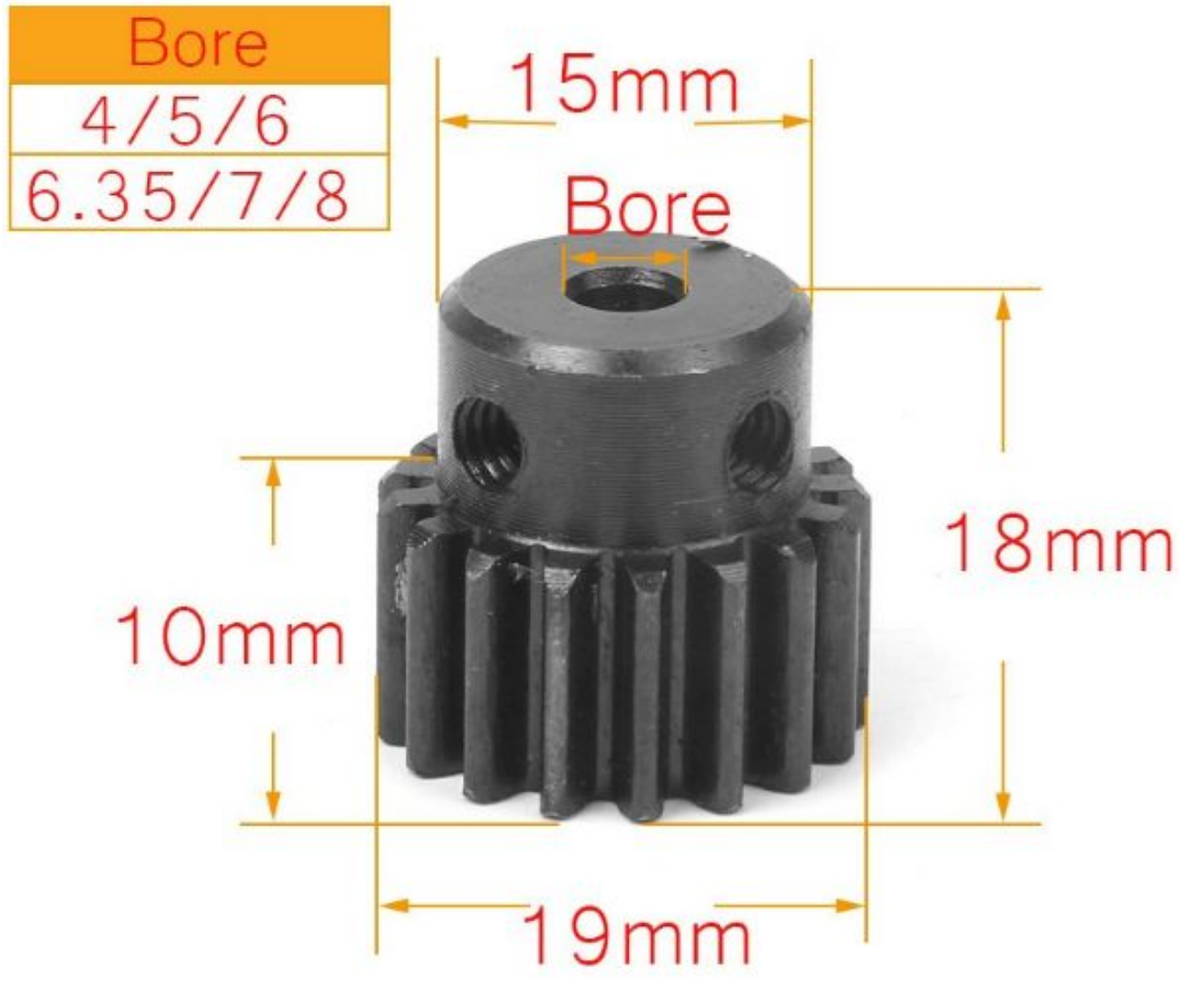}
\caption{The spur gear for steering and driving encoder. (Left) Steering motor encoder. The spur gear is affixed with a 7 mm gap from the casing. The arrow indicates the gap. (Center) Driving motor encoder. The spur gear is mounted without gap from the casing. (Right) The dimensions of the spur gear. 17 teeth spur gear and 7 mm bore size.}
\label{fig-spur-gear-assembled}
\end{figure}

Both driving and steering gearboxes are required to be modified. We examined each of the gearboxes and modified them to properly read the motor rotations. A 17 teeth spur gear was attached to each shaft of the encoders. The exact position of the spur gear in the encoder shaft is as follows. For the steering, the spur gear must be affixed with a 7 mm gap from the encoder outer casing. The spur gear is mounted, for the driving encoder, with no gap from the casing. See Fig.~\ref{fig-spur-gear-assembled} for more details.

\begin{figure}
\centering
\includegraphics[width=0.6\textwidth]{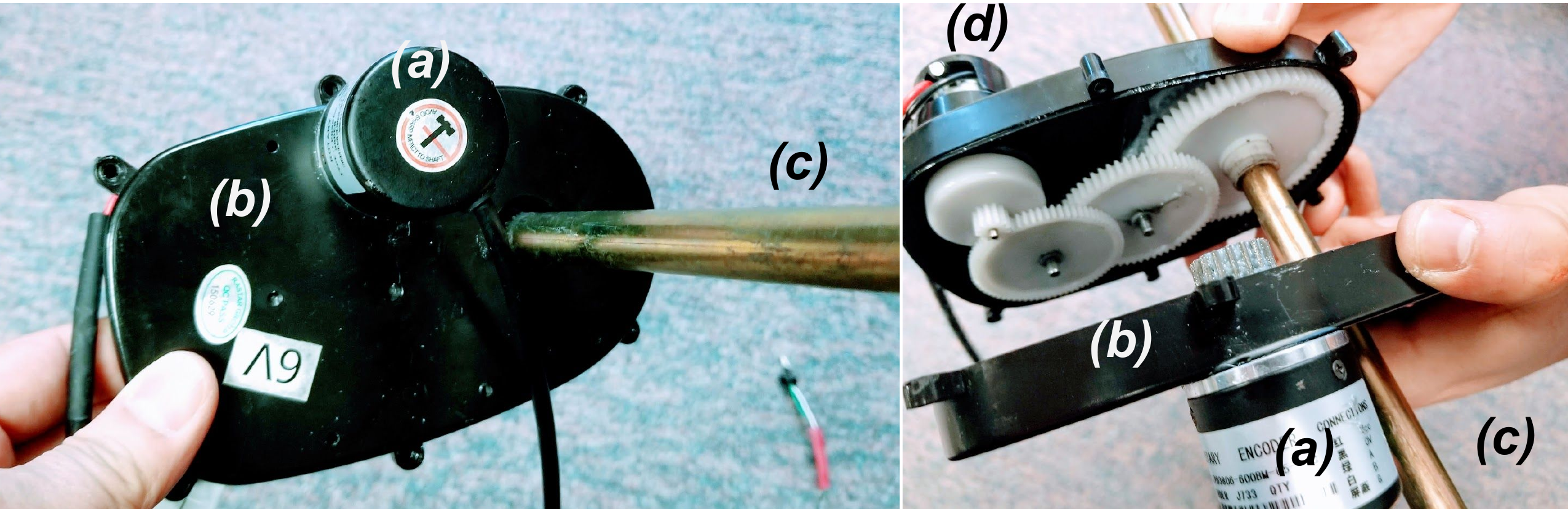}
\includegraphics[width=0.6\textwidth]{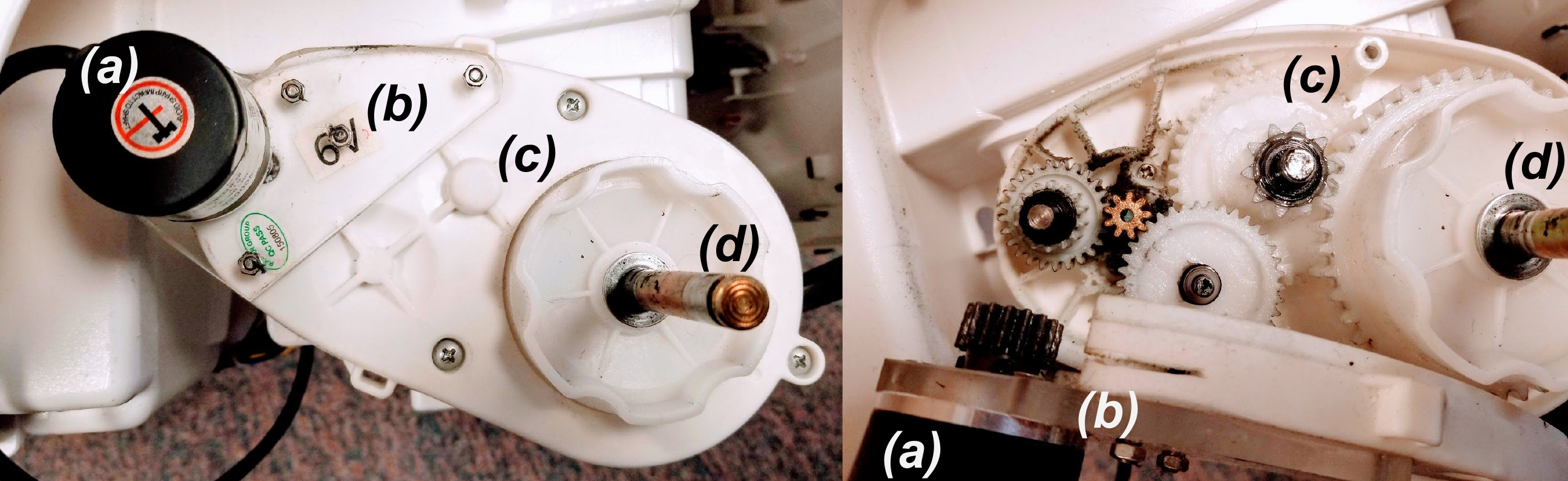}
\caption{Gearboxes. Top-Left: The steering wheel gearbox with an encoder attached. Top-Right: An exploded view of the modified steering gearbox. (a) The encoder. (b) The gearbox. (c) The steering wheel shaft. (d) The steering motor. Bottom-Left: The motor gearbox with an encoder attached. Bottom-Right: An exploded view of the modified driving gearbox. (a) The encoder. (b) The custom design support fixture. (c) The driving gearbox. (d) The driving motor shaft.
}
\label{fig-gearbox-open}
\end{figure}

\paragraph{Gearbox Modification}: The modified gearboxes are shown in Fig.~\ref{fig-gearbox-open}. These two gearboxes are in different positions in the vehicle. Due to the limitation of space, the position of the encoder gear’s teeth must be carefully chosen. The following sections have the details of it.

\begin{figure}
\centering
\includegraphics[width=0.65\textwidth]{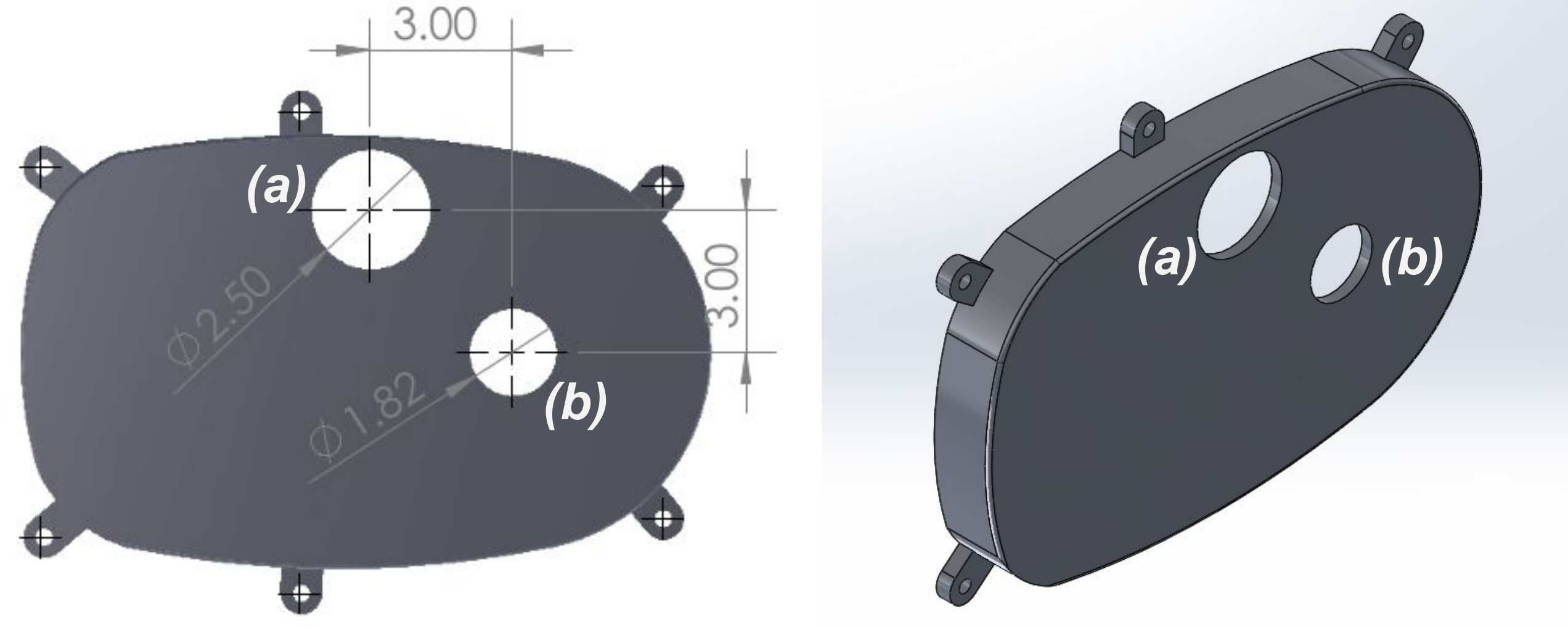}
\caption{The steering wheel gearbox. All units are in the metric system. We made a hole (a) in the top center for the encoder. The hole in the right (b) is for the steering wheel shaft. }
\label{fig-gearbox-steering-modification}
\end{figure}

The steering motor gearbox modification is relatively simple for the steering motor gearbox since there is enough room around the steering wheel shaft. We made a hole in the top center of the gearbox to affix the encoder to the largest gear in the gearbox. See Fig.~\ref{fig-gearbox-steering-modification} for more details. 

\begin{figure}
\centering
\includegraphics[width=0.65\textwidth]{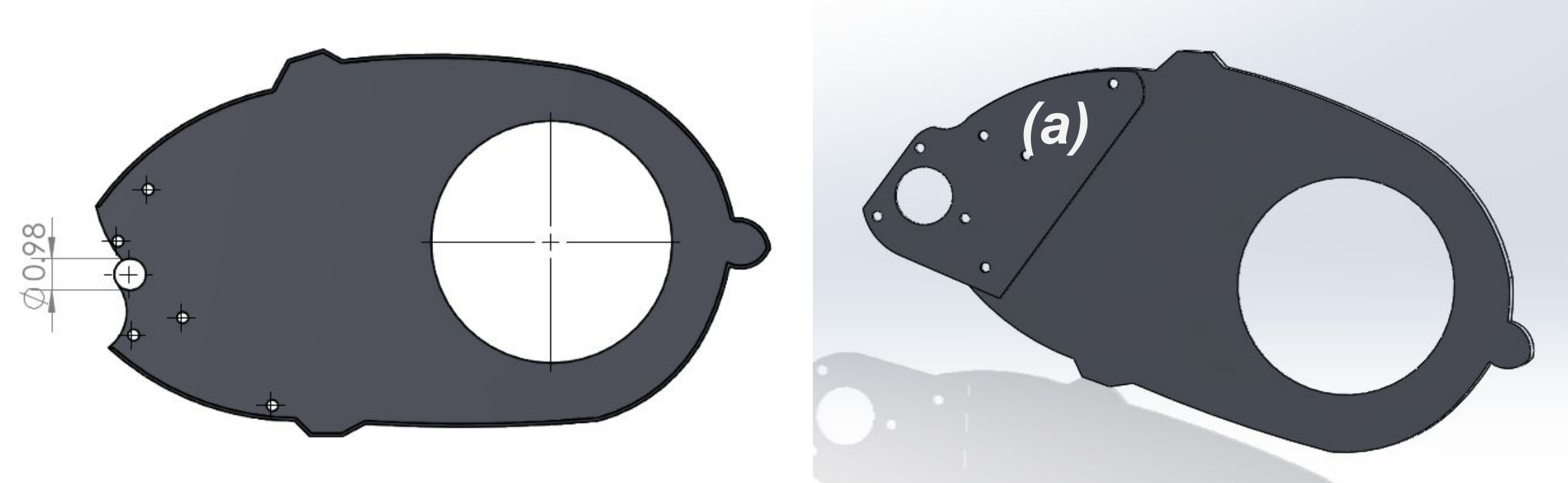}
\caption{The driving motor gearbox modification. All units are in the metric system. (a) The custom designed support fixture for the driving motor encoder. }
\label{fig-gearbox-drive-modification}
\end{figure}

\begin{figure}
\centering
\includegraphics[width=0.45\textwidth]{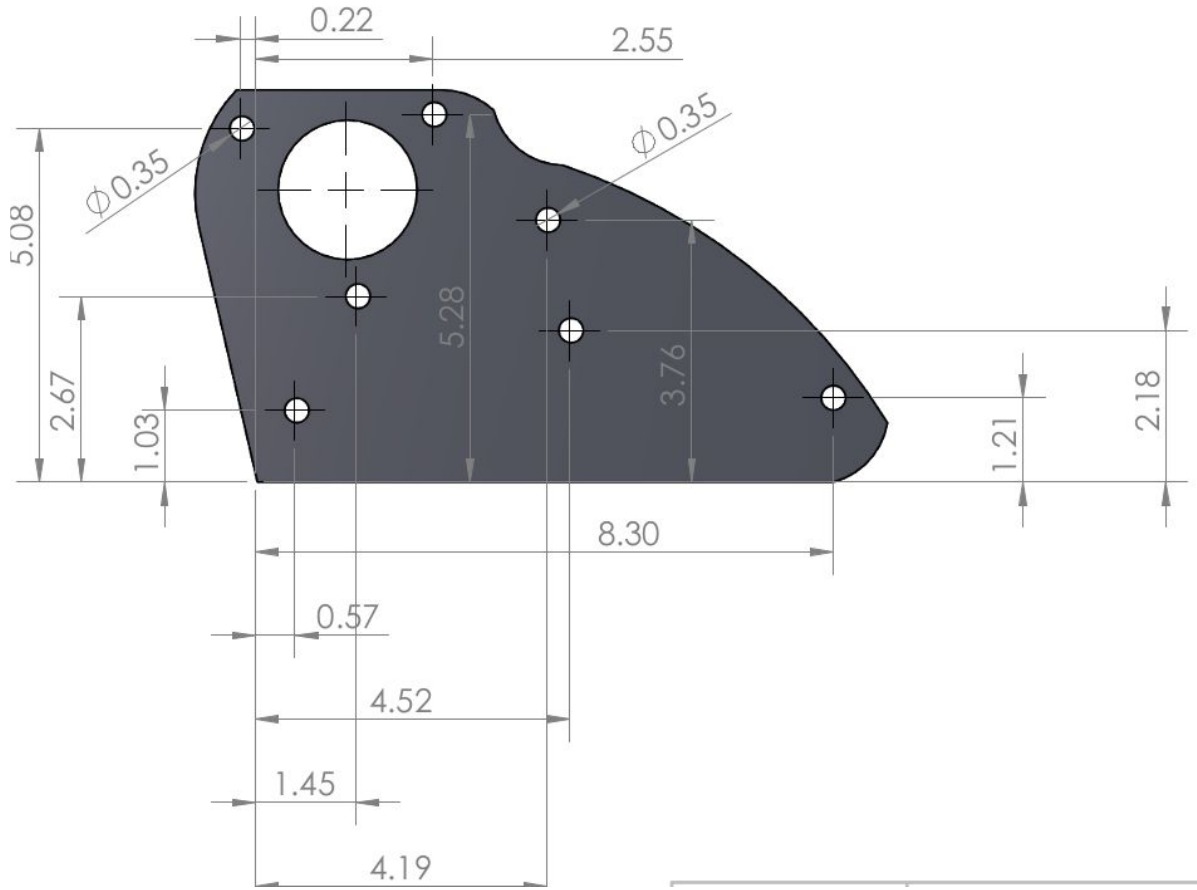}
\includegraphics[width=0.45\textwidth]{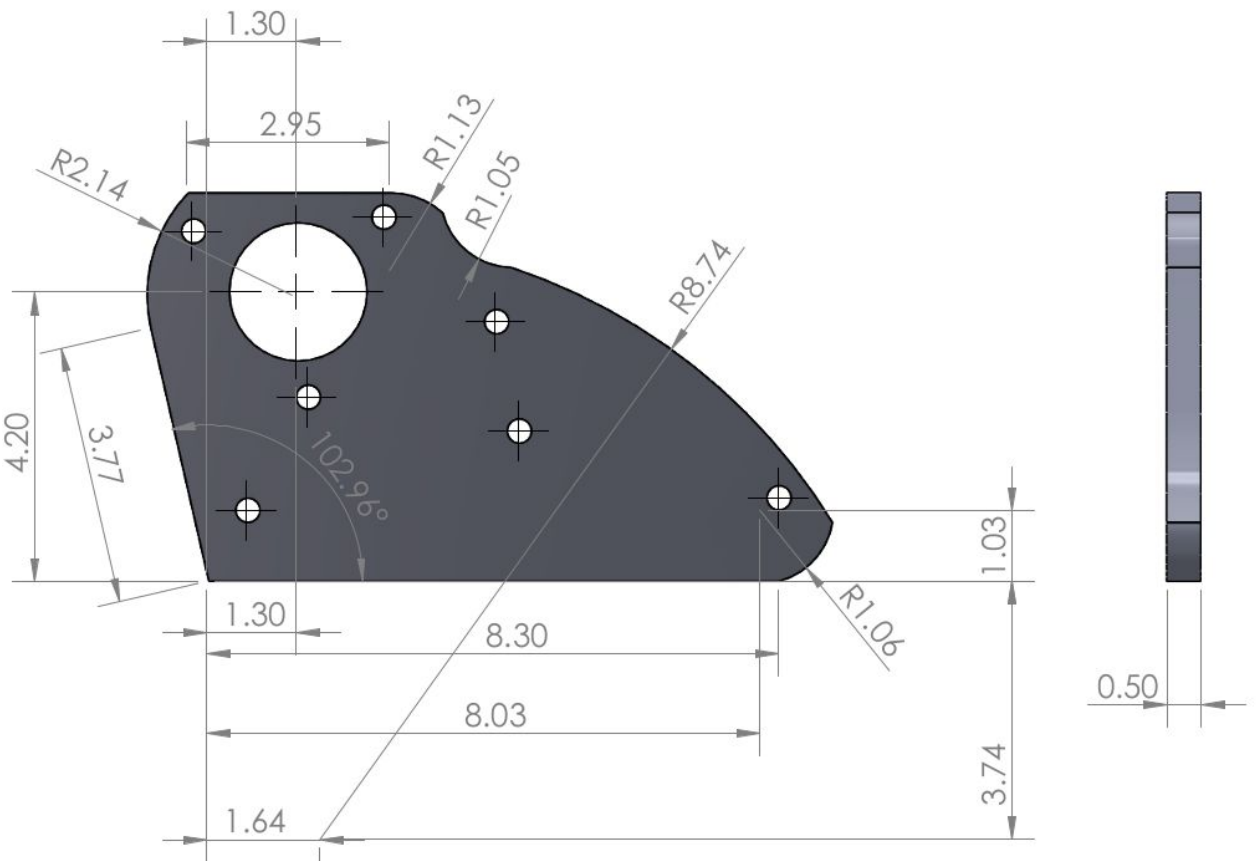}
\caption{The dimensions of driving gearbox support fixture. All units are in the metric system}
\label{fig-gearbox-drive-support-fixture}
\end{figure}

The driving motor gearbox needed a support structure. Due to the close distance between the driving wheel and gearbox, there was no space to place the encoder. A support fixture was designed and attached to the gearbox as shown in Fig.~\ref{fig-gearbox-drive-modification} and Fig.~\ref{fig-gearbox-drive-support-fixture}. A rear wheel is affixed to the hole via a shaft, and this blocks most of the gearbox. Not enough space is available to affix the motor encoder in the driving motor gearbox. Thus, we designed a fixture to hold the encoder that can be affixed to a gear inside the gearbox.


\subsubsection{Electrical Design}

\begin{figure}
\centering
\includegraphics[width=0.6\textwidth]{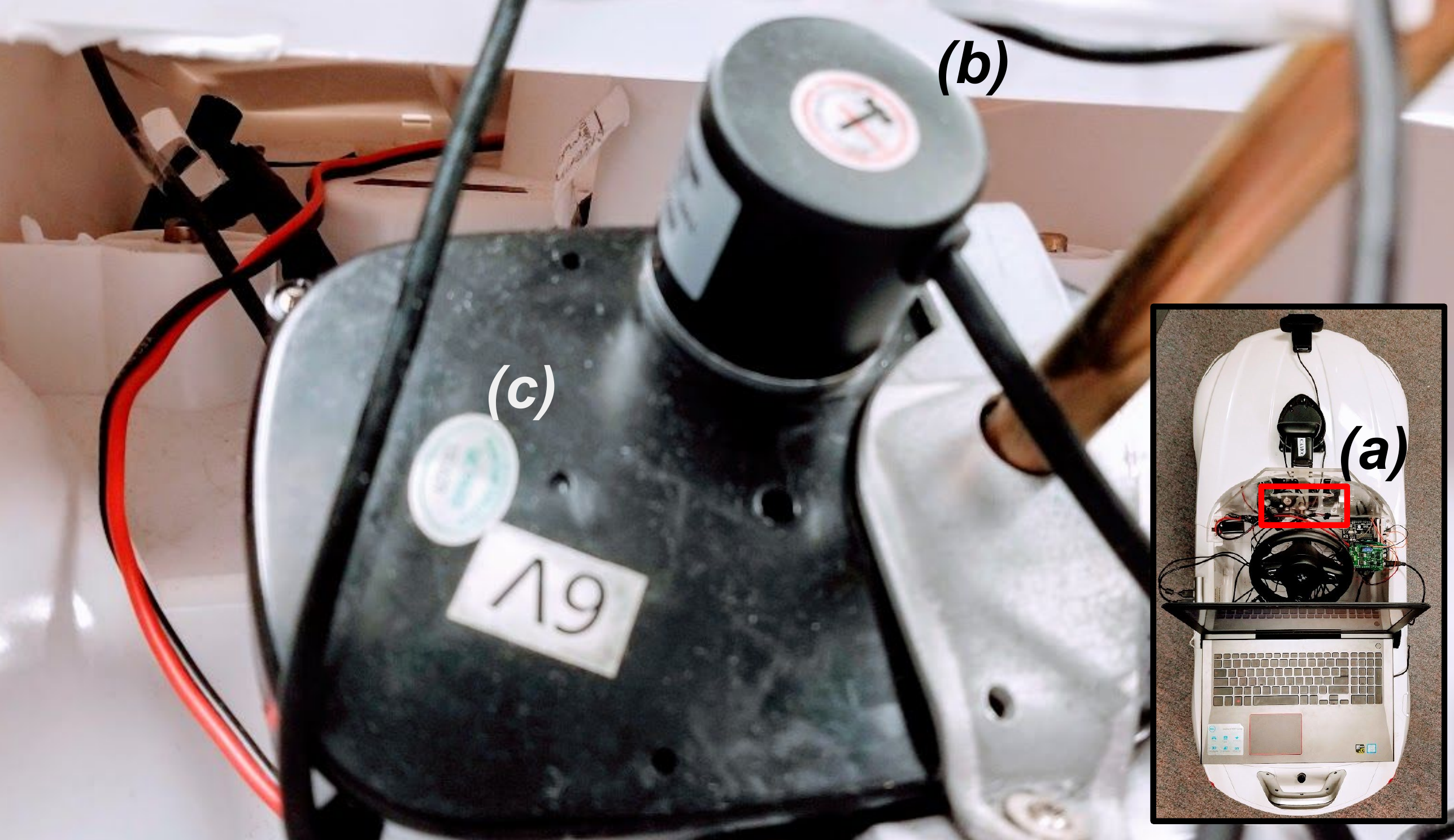}
\caption{The encoder attachment for the driving motor. The picture is the region (a) from the inset picture. (b) is the encoder and (c) is the gearbox in which the encoder is affixed. }
\label{fig-encoder-steering-attached}
\end{figure}

\begin{figure}
\centering
\includegraphics[width=0.6\textwidth]{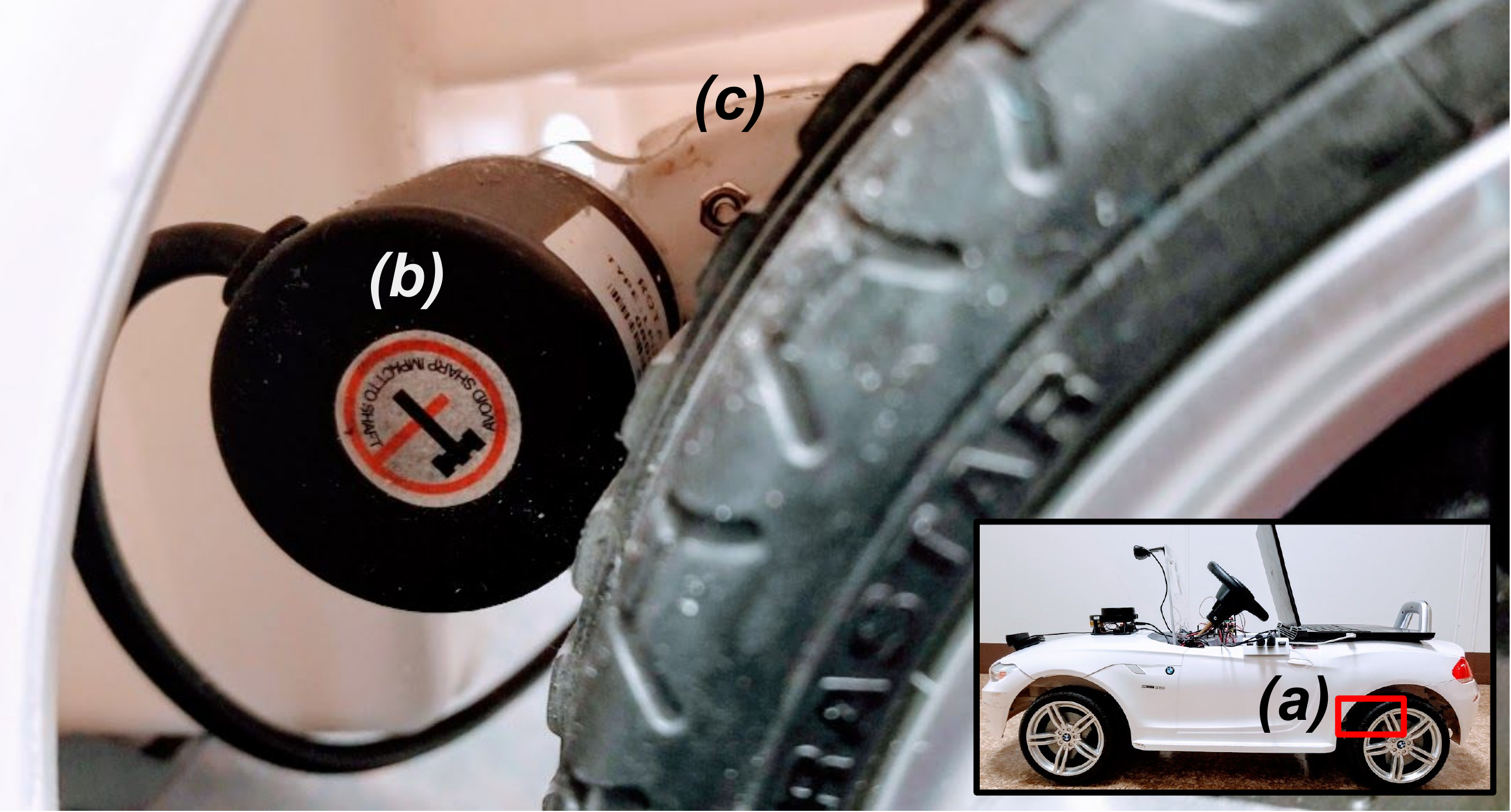}
\caption{The encoder attachment for the steering motor. The picture is the region (a) from the inset picture. (b) is the encoder and (c) is the gearbox in which the encoder is affixed. }
\label{fig-encoder-drive-attached}
\end{figure}

\paragraph{Encoders}: 
The ride-on-car is equipped with two electrical brushed motors. One motor is attached to the rear-left wheel and controls vehicle speed (Fig.~\ref{fig-encoder-drive-attached}), and the other attached to the steering wheel controls the direction of the vehicle (Fig.~\ref{fig-encoder-steering-attached}). To read the current speed and the direction of the car, two incremental encoders are attached to each of the motors. We used an incremental rotary encoder with a 6 mm shaft that has 600 pulses per revolution \citep{hn3806-ab-600n_hn3806-ab-600n_2016}. The size of the encoder is 38 x 35.5 mm and its shaft size is 6 x 13 mm. The size is small enough to be located in the gearbox of the driving motor and the shaft length is long enough to reach to the gears in the original gearbox. We implemented a quadrature decoder to convert the encoder signals into the direction and count. 


\begin{figure}
\centering
\includegraphics[width=0.6\textwidth]{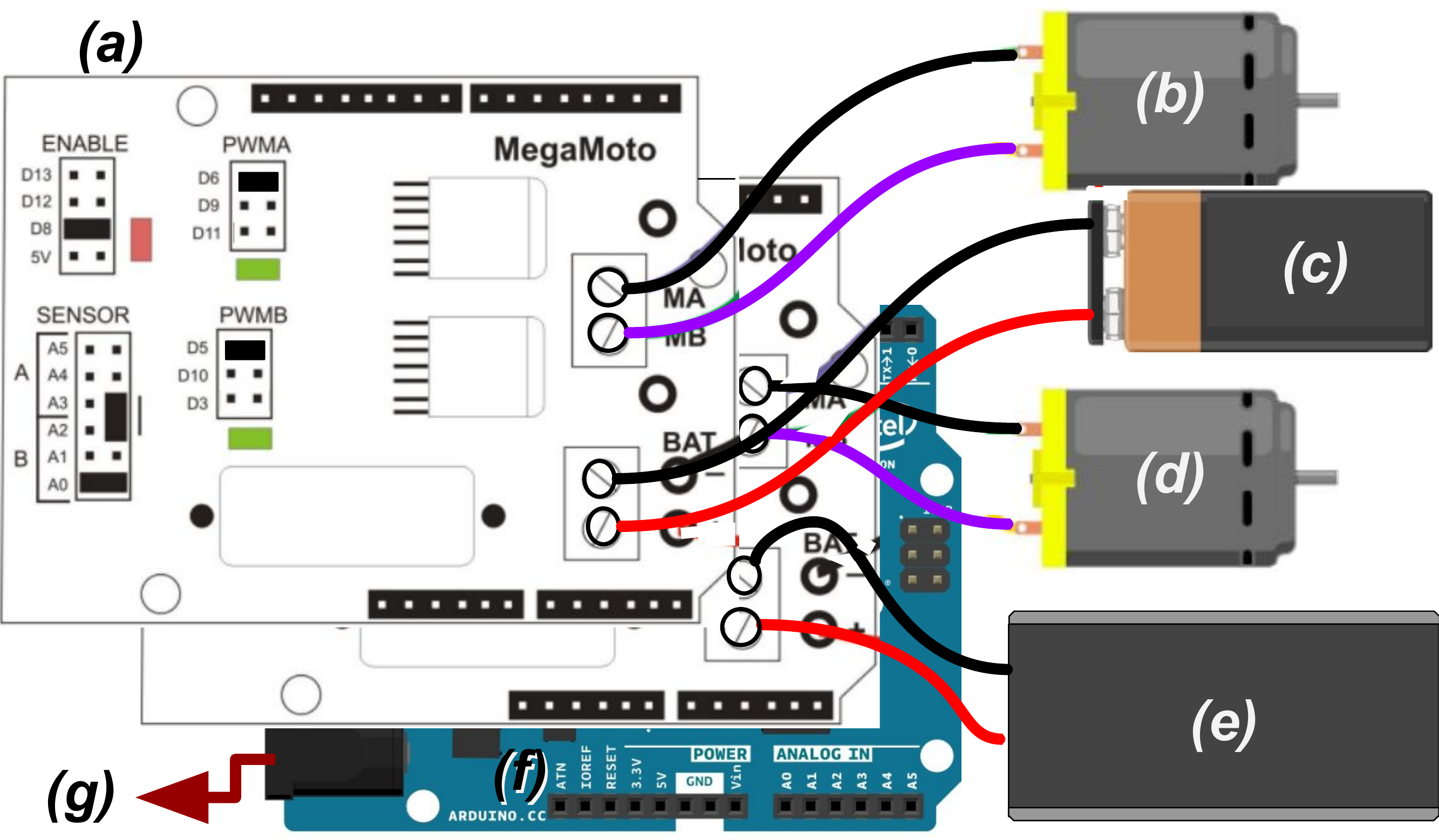}
\caption{The wiring diagram of motors and batteries. Both motor controllers are set as the MegaMoto H-bridge mode to drive DC brushed motors both forward and reverse. (a) The motor controller (b) The steering motor. (c) 9 V rechargeable battery. (d) The driving motor. (e) 9.6 V rechargeable battery. (f) The microcontroller board (g) The USB port to connect to the main laptop computer.}
\label{fig-wiring-motors}
\end{figure}

\paragraph{Microcontroller}:
Instead of using a high-performance microcontroller, we chose to use two Arduino Uno’s since it has a broad user community and simplicity of the use. Each of the microcontrollers is responsible for reading pulses from the incremental encoder since one microcontroller has only two interrupt pins that read Phase A and B signals from one encoder. These two microcontrollers are coupled via Inter-Integrated Circuit (I2C) in which the master/slave configuration is used. One I2C slave reads the encoder for the driving motor, and the other I2C master is responsible for not only reading the encoder for the steering wheel motor but also for sending PWM signals to control the speed and direction of the motors. For more details, see Fig.~\ref{fig-wiring-motors}.

\paragraph{Motors and Motor Controllers}:
This ride-on-car is equipped with two DC brushed motors (DC 6 V 25 W); one specified for controlling the platform steering angle and the other for controlling the speed of the platform. Two H-bridge motor controllers (MegaMoto Shield \citep{}) are used to deliver a continuous current (13 A). This motor controller is selected for its desirable characteristics of linear relationship between PWM signals and the output DC voltage. This linearity makes it easier to control the motors. 

\paragraph{Power and Wiring}:
The original vehicle has a control board to actuate motors. The wires from the motors must be connected to the MIR Vehicle’s motor control unit. The electric wiring diagram of motors and batteries are shown in Fig.~\ref{fig-wiring-motors}. 

\begin{figure}
\centering
\includegraphics[width=0.7\textwidth]{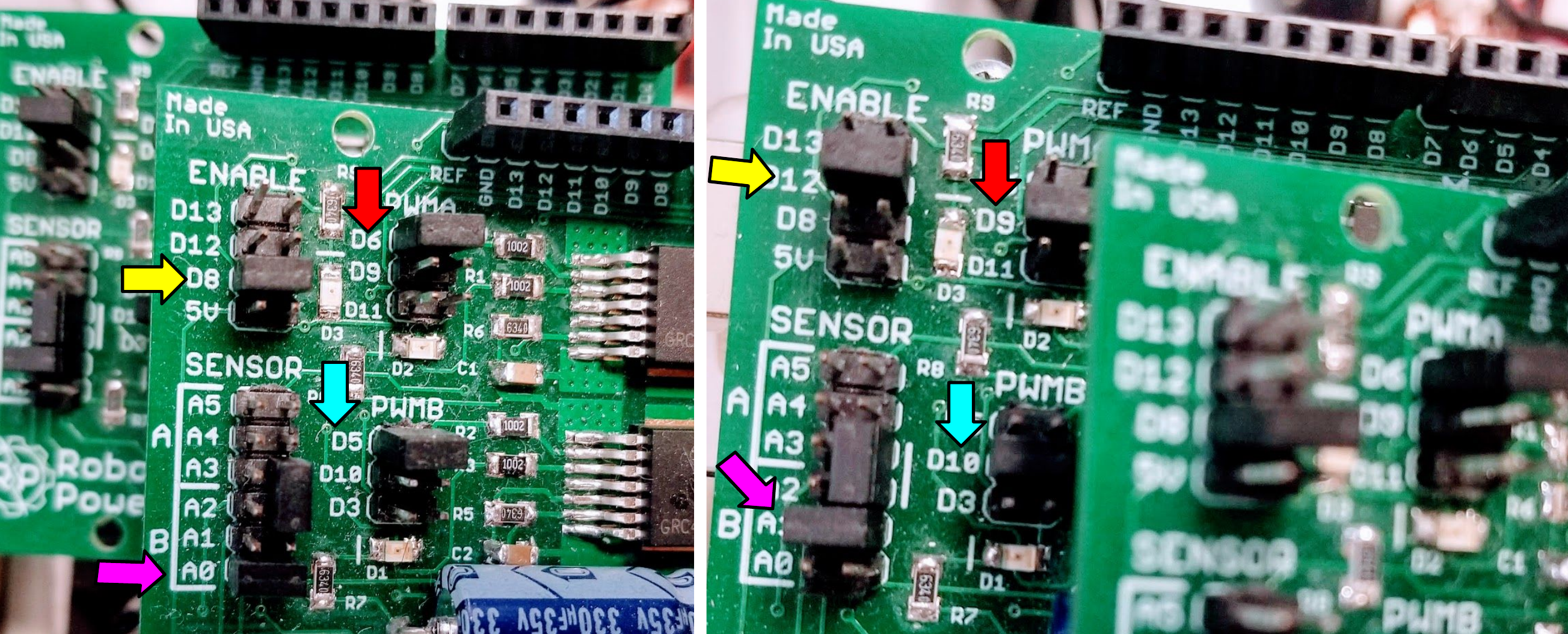}
\caption{The jumper settings for the motor controllers. Left: The steering motor controller. ENABLE (D8), SENSOR (A0), PWMA (D6), and PWMB (D5). Right: The driving motor controller. ENABLE (D12), SENSOR (A1), PWMA (D9), and PWMB (D10).}
\label{fig-motor-controller-jumpers}
\end{figure}

\begin{figure}
\centering
\includegraphics[width=0.65\textwidth]{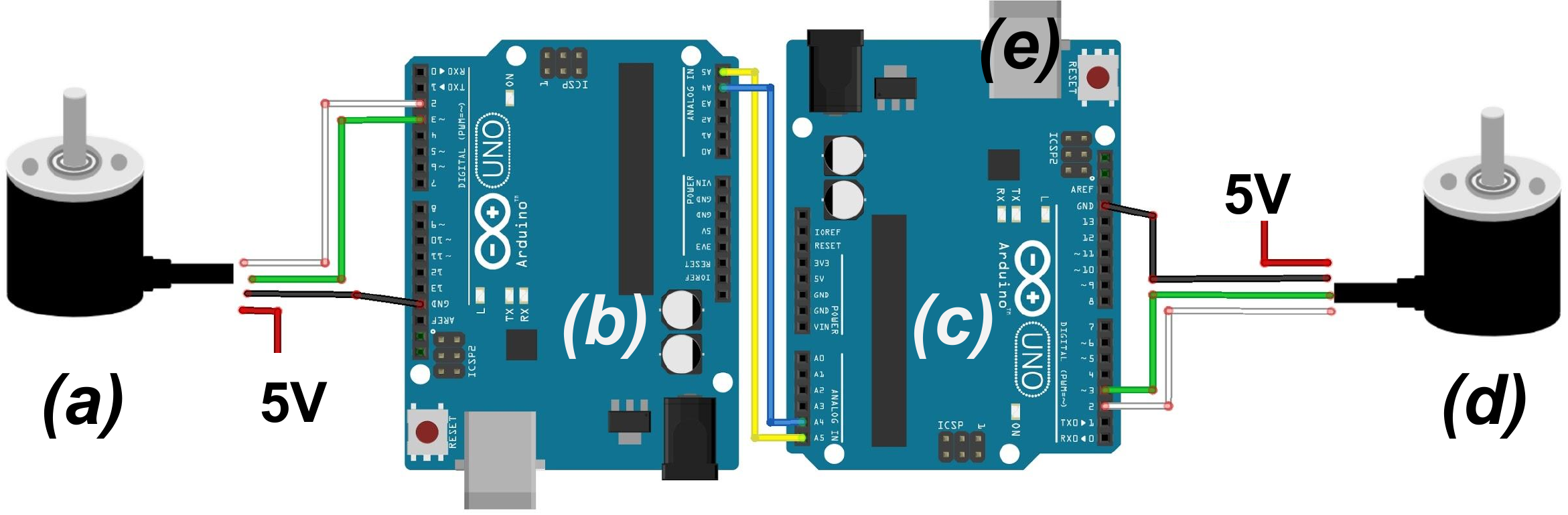}
\caption{The wiring diagram for motor encoders. (a) The driving motor encoder. (b) The I2C master. (c) The I2C slave that is connected to the main laptop computer. The motor controllers are connected to this microcontroller. (d) The steering motor encoder. (e) The USB connector that communicates with the laptop computer.}
\label{fig-wiring-encoders}
\end{figure}

Two MegaMoto H-bridge motor controllers are used to drive the DC brushed motors in both forward and reverse motions. We use a 9 V rechargeable battery for the steering motor and a 9.6 V (2 Ah) rechargeable battery for the driving motor. Each of the motor controllers is responsible for a specific motor. The jumpers on each board must be properly set. The details of the jumper settings are shown in Fig.~\ref{fig-motor-controller-jumpers}. Two motor encoders are used to read the driving motor speed and steering motor angle based on the motor revolution. One microcontroller is used to handle one motor encoder signals. So two microcontrollers are used and connected each other through the I2C interface to handle the two encoders. See Fig.~\ref{fig-wiring-encoders} for the details. The USB connector communicates with the laptop computer to send and receive data.


\subsection{Sensor Suite Design}

We tested two different RGB cameras and one RGBD camera for vision sensors. Two options of LIDAR sensors were analyzed and one was tested with the platform. An IMU is also an essential part of the vehicle to measure the vehicle platform’s movement and direction. We tested one 9 Degree of Freedom (DOF) IMU. All sensors we tested are integrated with the ROS.

\subsubsection{Camera Sensor}
Two USB cameras are used for evaluation purpose of the proposed system. More cameras can be attached to have rear views and/or side views. Our choice for a front view camera is Logitech C922X pro \citep{c922-specifications_c922_2016}. This camera has a high frame rate (60 fps in 720p) and wide horizontal and vertical Field of Views (FOV) (70.42 degrees and 43.3 degrees, respectively). The primary purpose of this front camera is to detect lane markings and other floor markings so that the proposed platform can be used to train a deep artificial neural network to clone a driver’s behavior. The second camera is Logitech C310 \citep{logitech-c310-hd-webcam_logitech_2019}. This camera was used to take videos from the car’s first-person perspective camera view.  

\subsubsection{RGBD Camera}
RGB with Depth (RGBD) cameras can be a viable option. We shortlisted RealSense depth cameras \citep{pruitt_choosing_2018},  D415 and D435. The major between the two is related to their specifications of the FOV and resolution. The D415 has a FOV or degrees with 1920 x 1080 pixels resolution, while the D435 has a FOV of 90 degrees with 1280 x 800 pixels resolution. 

\subsubsection{LIDAR}
LIDAR is one popular way to detect obstacles for ADAS and autonomous driving. We propose to use the Neato XV series laser distance sensor. There is no official vendor to sell this product since this sensor is being used for a robot vacuum cleaner. But this small 2D laser scanner is popular due to the affordability and ROS community supports compared to other small scale LIDAR products. The Neato XV laser can give five Hz in scanning with a range of five meters. Another viable option is to use YDLIDAR X4 \citep{ydlidar-x4_ydlidar_2016}. This gives the proposed platform enough scanning speed and range since it is used in indoor environment at low driving speed. YDLIDAR can scan in 6 ~ 12 Hz with up to around 10 meters range.

\subsubsection{IMU}
Razor 9 DOF IMU is a module with multiple sensors packed in one package \citep{sparkfun-9dof-razor-imu_sparkfun_2018}. The IMU is used to provide the platform with a sense of its linear acceleration, angular velocity, and magnetic field vectors. To calibrate the IMU module, each sensor (the accelerometer, magnetometer, and gyroscope) needs to be calibrated separately. The difference between the standard and the extended magnetometer is that the standard compensates only for the hard iron errors while the extended one compensates for hard and soft iron errors. The hard iron is created by the objects generating a magnetic field while the soft is the alterations in the existing magnetic field. To carry out the calibration process, Processing \citep{reas_environment_2001}, an Integrated Development Environment (IDE) for the electronic and visual design, is needed. It is recommended to mount the sensor on your platform to accommodate any magnetic field noise. For calibrating the extended magnetometer you may need to download a Processing library named EJML. These instructions are available at the beginning of the processing script  \citep{razor_imu_9dof_razor_imu_9dof_2018}.

\subsection{Software Design}

\subsubsection{Environment}
We use Ubuntu 16.04 LTS since the ROS Kinetic officially supports only Ubuntu 16.04 LTS. The ROS is not an operating system even if the name implies. It is middleware on top of an operating system (Ubuntu 16.04 LTS). The ROS provides device drivers, libraries, message-passing, and package management so that developers can create robot applications easily. Autonomous vehicles are basically intelligent robots that happen to have four wheels and steering. 

\subsubsection{Install ROS Packages}

The ROS packages for the sensors previously mentioned can be installed with the following commands.

\begin{verbatim}
$ sudo apt install ros-$ROS_DISTRO-usb_cam 
$ sudo apt install ros-$ROS_DISTRO-xv_11_laser_driver
$ sudo apt install ros-$ROS_DISTRO-razor-imu-9dof
$ sudo apt install ros-$ROS_DISTRO-joystick-drivers
\end{verbatim}

Some of the packages are not part of ROS Kinetic. For example, \verb|realsense2_cam| ROS package must be used for RealSense D415 and D435. The package, however, is not released with ROS Kinetic. The installation process is as follows. Clone the git repository from \citet{ros-wrapper-intel-realsense-devices_intelr_2019}, and copy the \verb|realsense-2.x.x| folder to \verb|mir_vehicle/src| to make it as a ROS package of the MIR-Vehicle workspace. To build this package, Intel RealSense SDK 2.0 must be installed. The details of the SDK installation can be found at \citet{intel-realsense-sdk-2.0_librealsense/distribution_linux.md_2019}. Two libraries (\verb|librealsense2-dkms| and \verb|librealsense2-utils|) must be installed, and one additional library, \verb|librealsense2-dev| must be installed to build the \verb|realsense2_cam| ROS package.
An available ROS package for the IMU is part of ROS Kinetic. If you use this package, there is no need to translate the coordinates. When you need to use the IMU in ROS with your own implementation, it is better to convert the roll, pitch, and yaw angles generated by the IMU to quaternions as to be in \verb|geometry_msgs/Quaternion| ROS message format \citep{quaternion_message_geometry_msgs/quaternion_2010}. The reason behind using quaternions is that two coordinates out of x, y, and z can be aligned. This is called gimbal lock that may occur and interfere with your readings. According to REP-103 \citep{rep-103_rep_2010}, ROS is using a right-handed coordinate system that uses x-axis for pointing forward and y-axis for pointing to left and z-axis for pointing up. Razor IMU, however, uses x-axis for pointing forward and y-axis pointing to right and z-axis pointing down. Therefore, necessary action needs to be taken to translate correctly the coordinate frames in your ROS node. 

\paragraph{Troubleshootings}:
Laser sensors are considered as a ttyACM or ttyUSB device. When you have the access denial error from the device, what you can do is as follows. Note that in case of \verb|ttyUSB|, use verb|ttyUSB| instead.

\begin{verbatim}
$ sudo chmod a+rw /dev/ttyACM<port_number>  
\end{verbatim}

What this does is to add read and write permission to the specified device.

\subsubsection{3D Vehicle Model}
\begin{figure}
\centering
\includegraphics[width=0.7\textwidth]{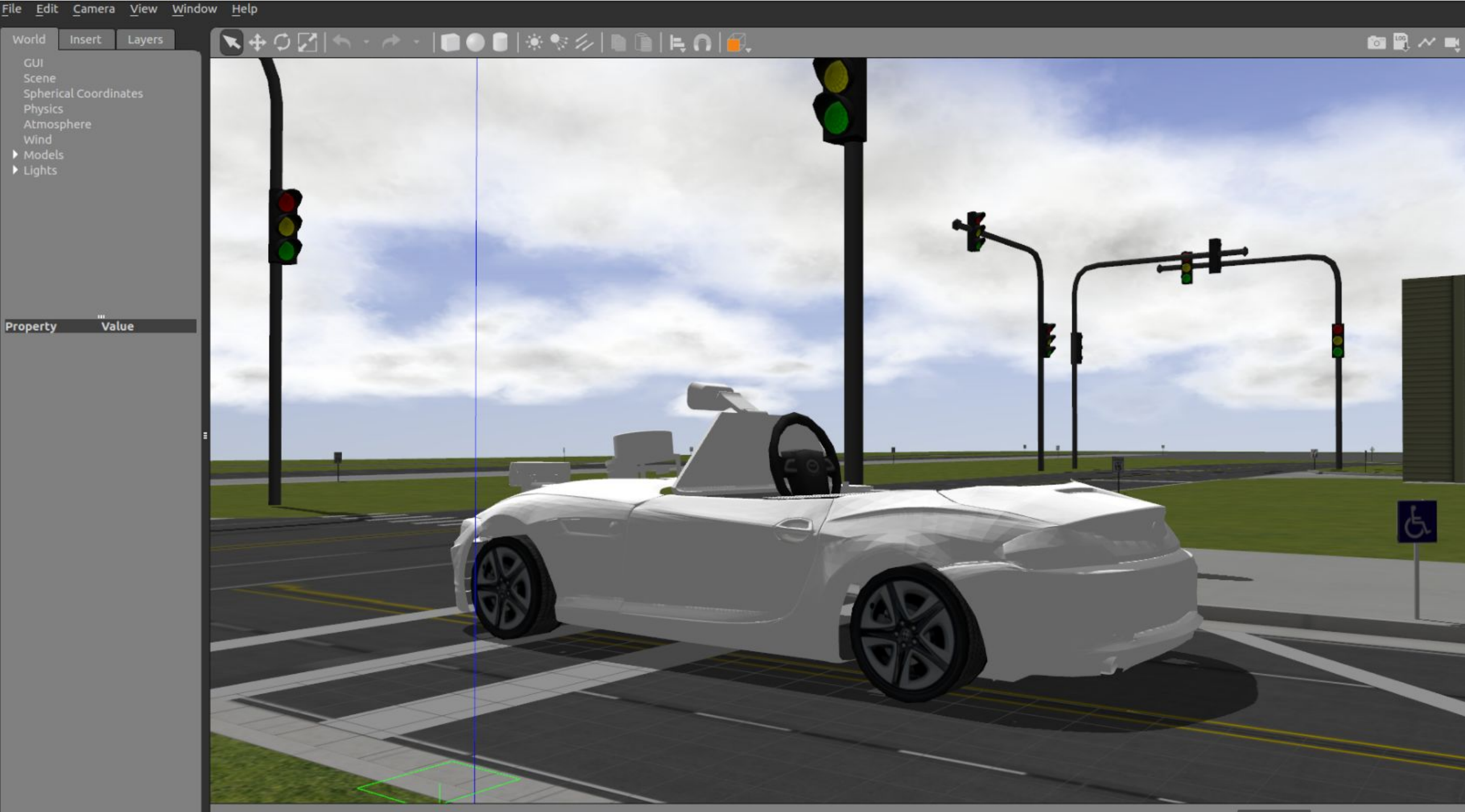}
\caption{A URDF model of the vehicle. The car model is deployed into a 3D simulated city.}
\label{fig-urdf-model-gazebo}
\end{figure}

We also created a Unified Robot Description Format (URDF) \citep{urdf_urdf_2019} model of the vehicle and sensor packages. This allows us to use the MIR Vehicle platform in a simulated environment. Fig.~\ref{fig-urdf-model-gazebo} shows that the MIR Vehicle is placed on a simulated environment.

\subsubsection{Microcontroller}

Two microcontrollers are used in the control unit. Each microcontroller supports up to two interrupt pins that are needed to read two channels from one encoder. One is for reading the driving motor encoder, and the other is not only for reading the steering motor encoder but runs a ROS node to communicate with the laptop. These two microcontrollers are connected via I2C interface. The ROS node is shown as \verb|serial_node| that publishes the \verb|encoder_pulse| topic and subscribes to the \verb|vehicle_control| topic. The \verb|joy2vehicle| node translates joystick control commands to the \verb|vehicle_control| topic that actuates the physical platform. The source code can be found at the \verb|arduino_driver| folder at \citet{kwon_mir_2019}.

\subsubsection{Laptop Computer}
A GPU powered laptop computer is the main computing source that runs the ROS core and complex perception and control algorithms based on Deep Learning libraries. Our recommendation for the GPU is GTX 1060 6 GB GDDR5 or above in terms of performance and GPU memory size. 

\paragraph{Remote Control}:
A 2.4 GHz wireless controller is used to send joystick commands that are translated to the vehicle control signals. We implemented a ROS node named joy2vehicle that can be found at the \verb|src/joy_control| folder at \citet{kwon_mir_2019}.  

\paragraph{Data acquisition}:
To develop autonomous vehicle applications, the vehicle platform must allow the users to collect data from sensors and the information from actuators. The MIR Vehicle platform offers the \verb|data_acquisition| package by which the users can collect images from cameras, distance measurements from a LIDAR, orientation and position from an IMU, and the speed and steering angles of the vehicle from motor encoders. The collected data location can be configured through \verb|rosparam|. We prepared a YAML file to configure that can be found at the \verb|src/data_acquisition/launch/| folder at \citet{kwon_mir_2019}. 

\subsubsection{Communication}
The motor control unit communicates with the laptop through \verb|rosserial|\citep{ros-serial_rosserial_2018} that is for interfacing with various microcontrollers through serial communication. Note that the ROS node inside the microcontroller is shown as \verb|/serial_node|. The ROS Master that is a computing platform running the ROS core is in the laptop computer where all other ROS nodes run. The ROS offers distributed computing architecture so no assumptions are made about the location of nodes and how they communicate with each other. 

\subsubsection{Set up ROS on the MIR-Vehicle}
\begin{figure}
\centering
\includegraphics[width=0.95\textwidth]{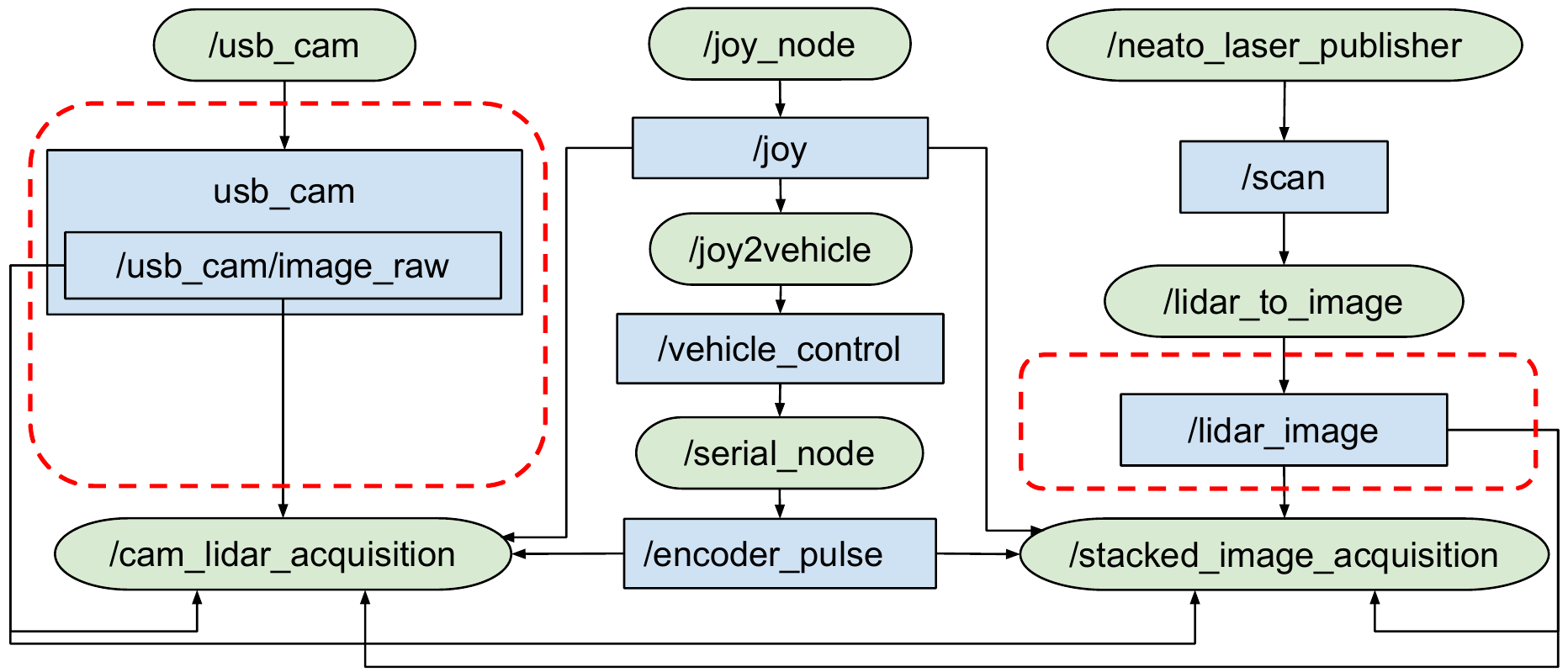}
\caption{ROS node graph for data acquisition. The ellipses are ROS node, and the rectangles are ROS topics. $/cam\_lidar\_acquisition$ and $/stacked\_image\_acquisition$ are examples of data acquisition nodes. The dotted red boxes are added when the data acquisition starts.}
\label{fig-node-graph-data-acquistion}
\end{figure}

An additional remote Git repository is prepared to install the ROS Kinetic with dependent packages \citep{kwon_ros_2019} for the proposed research platform.
Fig.~\ref{fig-node-graph-data-acquistion} shows the data acquisition node tree. Joystick commands are sent through the \verb|/joy| topic to the \verb|joy2vehicle| node. Then, the \verb|joy2vehicle| node translates the joystick commands to the MIR Vehicle’s control commands as a ROS topic named \verb|vehicle_control|. The \verb|/serial_node| subscribes to the \verb|/vehicle_control| topic and publishes the \verb|encoder_pulse| topic.  With the ROS graph (\verb|rqt_graph|), we can visualize how the nodes are communicating through ROS topics.    

\section{Results}
\begin{figure}
\centering
\includegraphics[width=0.53\textwidth]{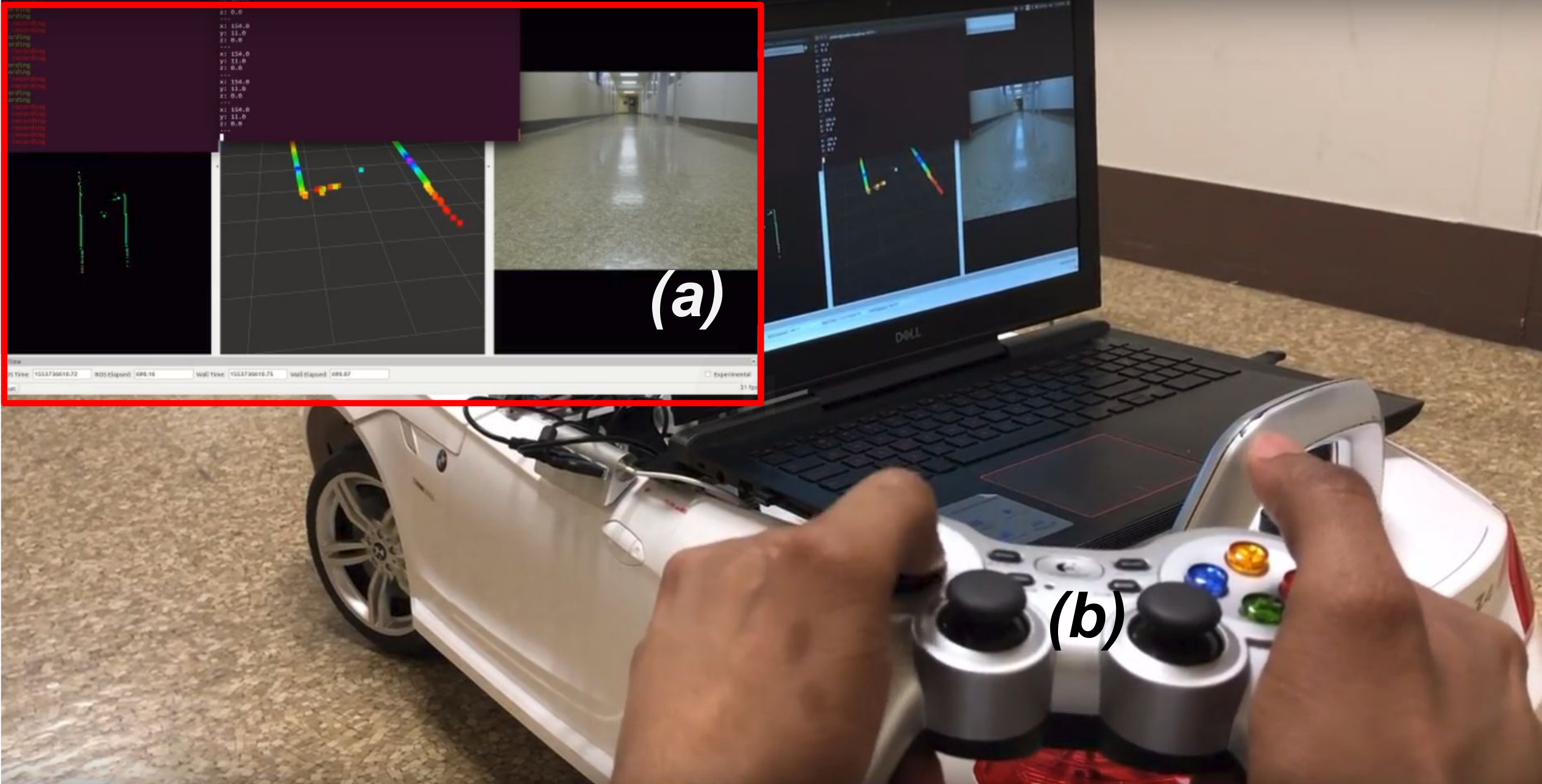}
\includegraphics[width=0.44\textwidth]{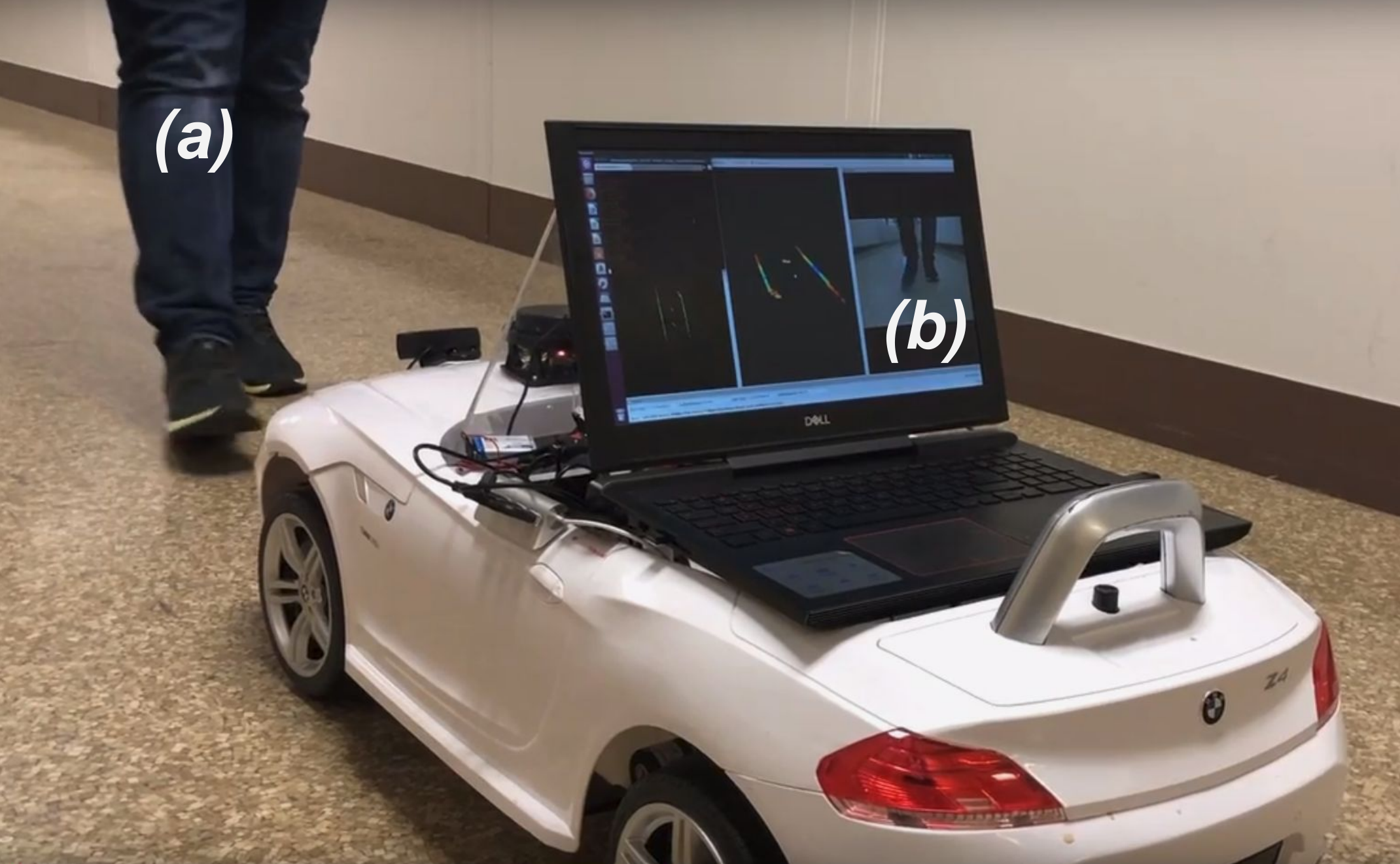}
\caption{Rviz screen with multiple ROS nodes to use sensor packages. Left: Remote control. (a) A screenshot of the ROS Visualization and our remote control ROS node. (b) A remote controller in action. Video: \url{https://youtu.be/LoJGX4yfP-A} Right: (a) An object. (b) ROS topic viewers from the LIDAR and the front camera. Video: \url{https://youtu.be/LoJGX4yfP-A} }
\label{fig-rviz-screen}
\end{figure}

The MIR Vehicle platform uses Python which is popular language for designing and testing perception and control algorithms in AI and data science community. ROS-based packages enable researchers to write applications in a modular form. Each ROS node (a computation unit) can easily communicate with others through the ROS framework that offers the distributed computing environment. Based on sensor data, the controllers produce commands that are translated to PWM signals for the actuators. The MIR Vehicle platform offers configuration files that can bring up all the necessary submodules to remote control the vehicle with active sensor packages. Also, the proposed research platform provides a data acquisition package that can be used to collect sensor data with steering wheel angle as well as driving motor speed. The datasets through the acquisition package are essential for behavior cloning approaches and sensor fusions for autonomous vehicle applications. All code for this paper is available at \citet{kwon_mir_2019}. 
For the evaluation of the system, experiments were developed and used to successfully control the vehicle using a remote controller, and collect images from a camera and distance measurements from a LIDAR. In Fig.~\ref{fig-rviz-screen}, we show the 3D visualization tool (RViz) when the MIR Vehicle platform receives commands from the joystick while displaying sensor data. 

\section{Conclusion}
This paper illustrates the design and transformation of a ride-on-car equipped with various sensors with an affordable budget for testing autonomous vehicle algorithms and/or ADAS systems. The platform designed here is scalable and economically viable. Sensors like LIDAR’s and cameras can be replaced according to the user\textsc{\char13}s requirements even though we proposed the optimal choices of sensors for this particular platform. Using hierarchical software design, the lower control part enabled by microcontrollers can be replaced with other feasible choices. Also, the higher computation part can be replaced with higher performance laptops. This would not be easy to accomplish if the platform uses embedded solutions such as NVIDIA’s Jetson TX1, or TX2 on a small chassis. The paper also emphasizes the use of the ROS with which we can not only pass messages between processes in different computing platforms but also visualize the messages both in numerical and visual forms using a visualization tool. We expect that the proposed platform can be widely used for research and education where various new algorithms can be tested and verified so that the wider user community can contribute to the ADAS/AD research areas.

\section{Appendix}
Table~\ref{tab-bom} is a Bill of Materials (BOM) of the proposed vehicle platform. The MIR Vehicle platform is flexible and scalable so that sensors can be easily replaced with others based on budget and specific requirements of the research activity from the users.

\begin{table}
\begin{tabular}{lllll}
\textbf{\textit{Item}}                                                                    & \textbf{\textit{Description}}                                                                  & \textbf{\textit{Q't}} & \textbf{\textit{Price}} & \textbf{\textit{Total}}    \\ 
\hline
Camera                                                                                    & Logitech C920                                                                                  & 1                     & \$66                    & \$66                       \\
Lidar Sensor                                                                              & Neato XV Lidar 5Hz scan rate                                                                   & 1                     & \$75                    & \$75                       \\
Arduino Uno                                                                               & 80 MHz frequency CPU\$                                                                         & 2                     & \$20                    & \$40                       \\
\begin{tabular}[c]{@{}l@{}}Incremental rotary \\encoder signswise \\LBD 3806\end{tabular} & 600 pulse per revolution                                                                       & 2                     & \$17                    & \$34                       \\
Other accessories                                                                         & Fixture, screws, and wires                                                                     & -                     & \$20                    & \$20                       \\
Logitech gamepad F710                                                                     & 2.4 GHz Wireless Controller                                                                    & 1                     & \$39                    & \$39                       \\
Motor Controller Shields                                                                  & Robot Power 13A 5-28V H-bridge                                                                 & 2                     & \$45                    & \$90                       \\
\begin{tabular}[c]{@{}l@{}}Newsmarts\\Spur Gear module\end{tabular}                       & 17 teeth 7mm bore -NS27IU001-25                                                                & 2                     & \$6                     & \$12                       \\
Extra Battery                                                                             & \begin{tabular}[c]{@{}l@{}}9.6V - 2000mAh \\Rechargeable with charger\end{tabular}             & 1                     & \$33                    & \$33                       \\
Ride-on-Car                                                                               & One Motor rear drive                                                                           & 1                     & \$210                   & \$210                      \\ 
\hline
\textbf{\textit{Total of the core items}}                                                 &                                                                                                &                       &                         & \textbf{\textit{\$619}}    \\ 
\hline
Optional Items                                                                            &                                                                                                &                       &                         &                            \\
Laptop                                                                                    & \begin{tabular}[c]{@{}l@{}}i7 7700HQ 2.8GHz 16GB RAM \\with GTX 1060 TI 6 GB GDDR\end{tabular} & 1                     & \$1,050                 & \$1,050                    \\
Intel Realsense                                                                           & Depth Camera                                                                                   & 1                     & \$149                   & \$149                      \\
IMU                                                                                       & Razor 9 DOF Sparkfun                                                                           & 1                     & \$36                    & \$36                       \\
Camera                                                                                    & Logitech C260                                                                                  & 1                     & \$19                    & \$19                       \\ 
\hline
\textbf{\textit{Total of optional items}}                                                 &                                                                                                &                       &                         & \textbf{\textit{\$1,254}}  \\
\hline
\end{tabular}
\caption{Bill of Materials}
\label{tab-bom}
\end{table}

\section*{Acknowledgment}
This work is partially supported by 2018 Faculty Research Fellowship of Kettering University, \textit{Development of Experimental Setup for Evaluating Vehicular Communication System (V2X)} and 2017 Provost’s Research Matching Fund of Kettering University, \textit{Development of Low-Cost Autonomous Vehicle Experimental Platform}.

\bibliographystyle{technopress} 
\bibliography{end/mir_vehicle}

\end{document}